\documentclass[10pt,twocolumn,letterpaper]{article}

\usepackage{cvpr}              
\definecolor{cvprblue}{rgb}{0.21,0.49,0.74}
\usepackage[pagebackref,breaklinks,colorlinks,allcolors=cvprblue]{hyperref}
\usepackage{stfloats}
\usepackage{capt-of}
\usepackage[most]{tcolorbox}

\usepackage{colortbl}
\usepackage{booktabs}
\usepackage{multirow}
\usepackage{graphicx}
\definecolor{lightblue}{RGB}{224, 235, 255}
\definecolor{lightgray}{RGB}{245, 245, 245}
\definecolor{lightyellow}{RGB}{255, 250, 205}
\usepackage{pgfplots}
\pgfplotsset{compat=1.18}

\usepackage{makecell}
\usepackage{pifont}
\usepackage{tikz}
\newcommand{\cmark}{\ding{51}}%
\newcommand{\xmark}{\ding{55}}%

\def\paperID{*****} 
\def\confName{CVPR}
\def\confYear{2026}

\title{CaST-Bench: Benchmarking \underline{Ca}usal Chain-Grounded \underline{S}patio-\underline{T}emporal \\Reasoning for Video Question Answering}

\author{Mingfang Zhang$^{*1,2}$,~~Jingjing Pan$^{*1}$,~~Ashutosh Kumar$^1$,~~Rajat Saini$^1$,~~Mustafa Erdogan$^1$,\\~~Hsuan-Kung Yang$^1$,~~Caixin Kang$^2$,~~Yifei Huang$^2$,~~Yoichi Sato$^2$,~~Quan Kong$^1$\\
$^1$Woven by Toyota,~~$^2$The University of Tokyo\\
{\tt\small \{firstname.lastname\}@woven.toyota,~~\{cxkang,hyf,ysato\}@iis.u-tokyo.ac.jp}
}

\begin{document}

\twocolumn[{
\begin{center}
\vspace{-2em}
  \maketitle
  \vspace{-1.5em}
  \includegraphics[width=\textwidth]{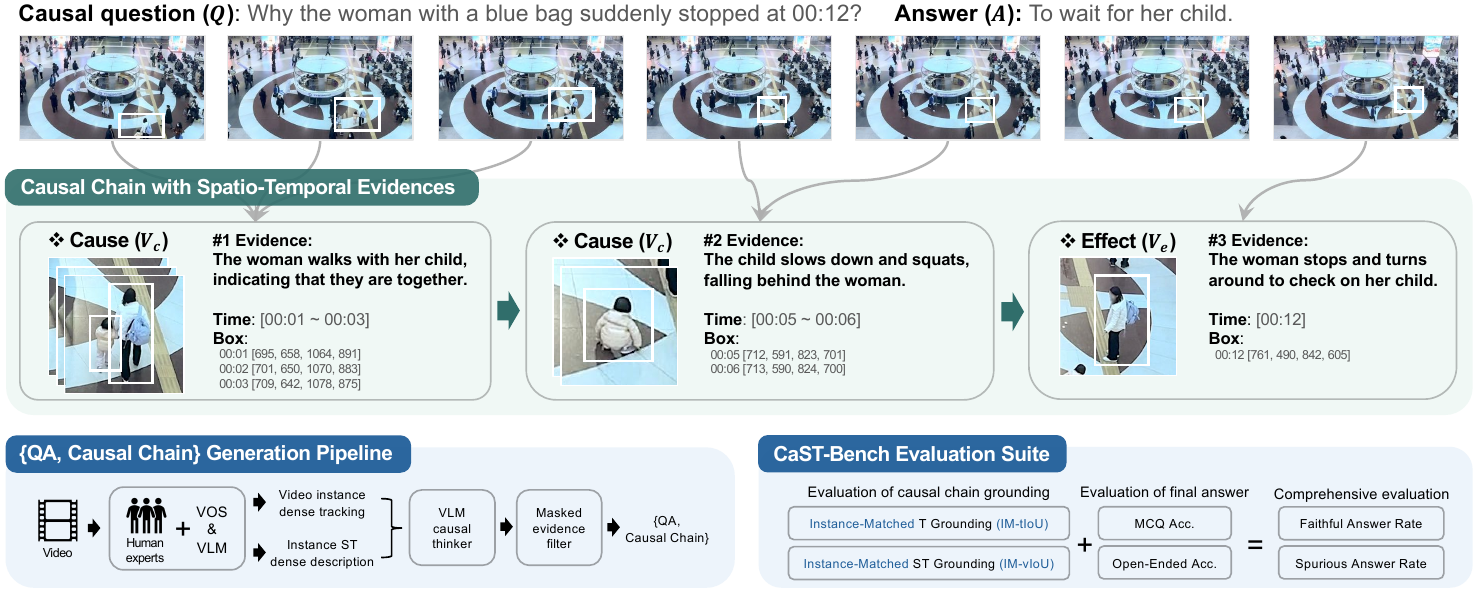}
  \captionof{figure}{\textbf{CaST-Bench Overview and Example Data.} Each QA in the benchmark is paired with a novel spatio-temporal (ST) causal chain. Unlike previous benchmarks, CaST-Bench requires models to actively search for both cause ($V_c$) and effect ($V_e$) evidences in order to construct a causal chain for question answering. 
  To excel on CaST-Bench, a model must produce an answer that is not only correct, but also faithfully grounded in an accurate ST causal chain.}
  \label{fig:overview}
\end{center}
}]
\begin{abstract}
\let\thefootnote\relax\footnotetext{ $^*$Joint first authors}
Cause-and-effect reasoning in video is a significant challenge for Vision-Language Models (VLMs), as it requires going beyond surface-level perception to a deeper understanding of causal mechanisms. 
However, existing benchmarks rarely provide the fine-grained, grounded evidence needed to rigorously evaluate this capability. 
To address this gap, we introduce CaST-Bench, a benchmark for \underline{Ca}usal Chain-Grounded \underline{S}patio-\underline{T}emporal Video Reasoning.
CaST-Bench presents complex causal questions that require models to identify and localize a chain of multiple spatio-temporal evidences.
Through a human-AI collaborative pipeline, we construct a high-quality dataset of 2,066 questions over 1,015 videos, with causal chains annotated by temporal segments and bounding-box tracks.
Furthermore, we design a comprehensive evaluation suite with novel metrics that assess not only answer correctness but also the capability for visual evidence grounded reasoning.
This grounding is crucial for improving accuracy by mitigating spurious correlations and for enhancing user trust by making models more transparent.
Our experiments show that current VLMs struggle with causal questions, largely due to their limited ability to construct precise and grounded causal chains. This highlights 
an important direction for improving future VLMs.
Homepage: \url{https://woven-by-toyota.github.io/CaST-Bench}.

\vspace{-1.5em}
\end{abstract}  

\section{Introduction}
\label{sec:intro}

The rapid progress of Vision-Language Models (VLMs) has greatly advanced video understanding, but reasoning about cause-and-effect in video events remains a significant challenge. Causal reasoning \cite{pearl2009causal, hernan2010causal} is critical for tasks such as video analysis, explanation, and anticipation, where understanding \textit{why} events occur and \textit{how} they influence each other goes beyond surface-level visual perception.

\begin{figure}
    \centering
    \includegraphics[width=0.99\linewidth]{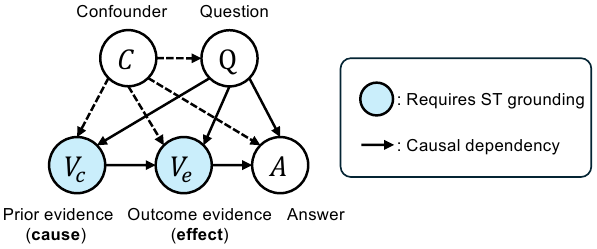}
    \caption{\textbf{Causal Graph for CaST-Bench Task Definition.} Given a video $V$ and a question $Q$, the task is to discover and ground both the prior evidence (cause) $V_c$ and the outcome evidence (effect) $V_e$ spatially and temporally to form a causal chain that leads to the answer $A$. Confounder factors $C$ may introduce spurious correlations that make the task challenging.}
    \vspace{-1em}
    \label{fig:causal_graph}
\end{figure}

Most existing video understanding benchmarks \cite{li2024mvbench, zhang2020videostg} focus on descriptive questions, such as ``what color is the woman's bag?'', which require models to \textit{directly locate} and describe objects and actions. In contrast, causal reasoning demands a deeper understanding of why events happen, e.g., ``why did the woman stop?'' (as shown in \cref{fig:overview}). 
This requires models to \textit{actively search} for a chain of causally linked evidences, including both the cause ($V_c$: ``The child bent down.'') and the effect ($V_e$: ``The woman waited.'') as conceptually illustrated in \cref{fig:causal_graph}. Notably, some key evidences may not be mentioned in the question.

This naturally raises an important question: Can current VLMs accurately localize these linked evidences in videos to answer causal questions?
Addressing this challenge is fundamental to the advancement of future VLMs, as accurate evidence grounding yields two key benefits. First, it allows models to base their reasoning on genuine causal cues rather than spurious linguistic or visual correlations (the ``confounder'' in \cref{fig:causal_graph}), leading to improved accuracy. Second, presenting visual evidence alongside answers makes VLMs more transparent and trustworthy to VLM users.

However, most existing video benchmarks \cite{li2024mvbench, fu2025videomme} lack visual grounding in the reasoning process. Some \cite{chen2024cgbench, xiao2024nextgqa} focus solely on temporal grounding and do not capture spatial ambiguity in cluttered scenes (\cref{fig:overview}). Others \cite{lei2020tvqa+, cheng2025vstarbenchmark} provide grounding only for objects mentioned in the question while overlooking the causally linked evidence necessary for genuine causal reasoning.

In this work, we introduce CaST-Bench for a novel task of \underline{Ca}usal Chain-Grounded \underline{S}patio-\underline{T}emporal Reasoning for Video QA. It differs from existing benchmarks in several key aspects. 1) \textbf{Causal Reasoning Focus}: CaST-Bench presents causal questions that require models to actively search for, integrate, and analyze multiple hidden clues to construct a complete causal reasoning chain. 2) \textbf{Fine-Grained Spatio-Temporal (ST) Evidence}: CaST-Bench provides detailed spatio-temporal evidences for each causal QA pair. This includes explicit timestamps and tracked bounding boxes that localize the key visual evidence responsible for the causal relationship.


We construct CaST-Bench from open-domain videos depicting complex scenes. To support fine-grained causal reasoning, we design a human–AI collaborative pipeline that first produces dense, instance-level descriptions and then generates causally grounded QA pairs based on spatio-temporal evidence chains. 
Our pipeline features human-verified, timestamp-accurate evidence annotations and applies mask-based evidence validation to ensure the completeness of the causal chain we provide. The final benchmark includes 2,066 challenging questions in 1,015 videos, providing a rigorous basis for evaluating causal understanding in videos.


To evaluate performance on CaST-Bench, we introduce a novel evaluation suite that measures not only answer correctness, but also the ability to provide visually grounded reasoning. Across 15 representative VLMs, performance on causal questions remains far below human ability. Human annotators reach 91.89\% accuracy on CaST-Bench, while the best proprietary model Gemini-2.5-Pro~\cite{comanici2025gemini} scores 50.34\% and the best open-source model Qwen3-VL~\cite{qwen3vl2025} scores 45.30\%.
Our ablation studies further indicate that constructing precise, grounded causal chains is key to improving answer accuracy and point to promising directions for future research. In summary, our main contributions are as follows:

\begin{itemize}
  \item We introduce Causal Chain-Grounded Spatio-Temporal Reasoning for Video QA, a new task designed to push models beyond perception to causal reasoning grounded in fine-grained ST evidences.
  
  \item We propose CaST-Bench, a benchmark that emphasizes causal reasoning and ST evidence grounding, along with an evaluation suite purpose-built for this task.
  
  \item Our experimental results indicate that current models struggle with causal questions, and we demonstrate that incorporating accurately grounded causal chains in the reasoning process is a crucial step toward improving their answer accuracy.
\end{itemize}

\section{Related Works}
\label{sec:rw}


\paragraph{Reasoning Vision-Language Models} Vision-language models (VLMs) \cite{zhang2024llavanextvideo, li2023videochat, maaz2024video, li2024llama, song2024moviechat, chen2024internvl25, Qwen2.5-VL} have progressed rapidly from basic perception to sophisticated reasoning. Techniques such as Chain-of-Thought (CoT) and reinforcement learning \cite{guo2025deepseek, li2025videochat, feng2025video, zhang2025tinyllava, chen2025exploring, wang2025time} further strengthen their ability to tackle complex video tasks. Recently, some image reasoning models \cite{zheng2025deepeyes, xu2025visual, zhang2025chain, lai2025mini, wang2025traceable} have introduced spatial grounding to improve credibility and attention to fine-grained details, while some video reasoning models \cite{meng2025openo3video, shi2024agentdistill, zhang2025thinking, su2025pixel, chen2025cross, zang2023discovering} apply temporal grounding to better capture event order. However, joint spatio–temporal grounding remains rare, and causal linkage across visual evidences is often overlooked.

\begin{table*}[!t]
\centering
\small
\renewcommand{\arraystretch}{1.2}
\setlength{\tabcolsep}{10pt} 
\begin{tabular}{lcccccccc}
\hline
\makecell[c]{\textbf{Video}\\\textbf{Benchmarks}} &
\makecell[c]{\textbf{Video}\\\textbf{Domain}} &
\textbf{Videos} &
\textbf{Anno.} &
\makecell[c]{\textbf{QA}\\\textbf{Pairs}} &
\makecell[c]{\textbf{QA}\\\textbf{Types}} &
\makecell[c]{\footnotesize\textbf{Direct}\\\footnotesize\textbf{Spatio-Temporal}\\\footnotesize\textbf{Localization}} &
\makecell[c]{\footnotesize\textbf{Chained}\\\footnotesize\textbf{Spatio-Temporal}\\\footnotesize\textbf{Evidences}} \\
\hline
MVBench \cite{li2024mvbench} & open & 3641 & A\&M & 4000 & MCQ & {\color{red}\xmark} & {\color{red}\xmark} \\
Video-MME \cite{fu2025videomme} & open & 900 & M & 2700 & MCQ & {\color{red}\xmark} & {\color{red}\xmark} \\
TVQA+ \cite{lei2020tvqa+} & TV show & 403 & M & 2821 & MCQ & {\color{green}\cmark} & {\color{red}\xmark} \\
V-STaR \cite{cheng2025vstarbenchmark} & open & 743 & A\&M & 2090 & OE & {\color{green}\cmark} & {\color{red}\xmark} \\
EgoExoBench \cite{he2025egoexobench} & open & 29083 & M & 7300 & MCQ & {\color{green}\cmark} & {\color{red}\xmark} \\
CG-Bench \cite{chen2024cgbench} & open & 1219 & M & 12129 & MCQ \& OE & {\hspace{0.36em}\color{red}\xmark}* & {\color{red}\xmark} \\
VR-Bench \cite{yu2025vrbench} & narrative & 960 & M & 8243 & MCQ \& OE & {\color{red}\xmark} & {\hspace{0.36em}\color{red}\xmark}* \\
Video-Holmes \cite{cheng2025videoholmes} & short film & 270 & A\&M & 1837 & MCQ \& OE & {\color{red}\xmark} & {\hspace{0.36em}\color{red}\xmark}* \\
Causal-VidQA \cite{li2022causalvidqa} & action & 5429 & A\&M & 21716 & MCQ & {\color{red}\xmark} & {\hspace{0.36em}\color{red}\xmark}* \\
NExT-GQA \cite{xiao2024nextgqa} & open & 990 & M & 5553 & MCQ & {\hspace{0.36em}\color{red}\xmark}* & {\hspace{0.36em}\color{red}\xmark}* \\
CaST-Bench (Ours) & open & 1015 & A\&M & 2066 & MCQ \& OE & {\color{green}\cmark} & {\color{green}\cmark} \\
\hline
\end{tabular}%
\caption{\textbf{Comparison Between CaST-Bench and Existing Video QA Benchmarks.}
A and M denote automatic and manual annotations; MCQ and OE denote multiple-choice and open-ended questions. The marker (*) means support for \textit{either} spatial (S) \textit{or} temporal (T) annotations, but not both.
While some benchmarks include ST grounding task, they focus on direct ST localization rather than active discovering chained ST evidences. \cite{yu2025vrbench, cheng2025videoholmes} include T evidences but they don't evaluate grounding accuracy. CaST-Bench is the first to support both spatial and temporal chained evidence grounded causal reasoning and comprehensive evaluation.}
\label{tab:benchmark_comparison}
\vspace{-1em}
\end{table*}


\paragraph{Video Benchmarks for VLMs} Video understanding benchmarks have expanded rapidly to evaluate the growing capabilities of VLMs. Comprehensive video benchmarks like \cite{li2024mvbench, fu2025videomme, xu2017video, xiao2021next}
have been introduced to evaluate temporal reasoning and long-form video understanding. To assess the credibility of VLMs, some benchmarks \cite{lei2020tvqa+, cheng2025vstarbenchmark, he2025egoexobench, chen2024cgbench, zhang2025video, wang2024videocot} annotate timestamps and bounding boxes for queried objects, enabling evaluation of \textit{direct spatio-temporal localization} (as shown in \cref{tab:benchmark_comparison}). Meanwhile, several recent benchmarks \cite{yu2025vrbench, cheng2025videoholmes, li2022causalvidqa, xiao2024nextgqa} aim to measure advanced causal reasoning abilities that require model to actively search for a chain of causal evidences, but they either provide solely narrative textual clues \cite{li2022causalvidqa, foss2025causalvqa} or lack spatio-temporal annotations \cite{yu2025vrbench, cheng2025videoholmes, xiao2024nextgqa, chen2024mecd}. In contrast, our CaST-Bench fills this gap as the first benchmark to provide \textit{chained spatio-temporal evidences} (as shown in \cref{tab:benchmark_comparison}), and a comprehensive evaluation suite that enables rigorous evaluation of causal reasoning process.
\section{Task Definition}

We introduce Spatio-Temporal Causal Chain-Grounded Video QA, a new task designed to push models beyond perception to causal reasoning grounded in fine-grained spatio-temporal evidences. The task input is a video $V$ and a question $Q$. The model must output a final answer $A$ after a variable-length list of evidences $E_C = [E_1, \dots, E_K]$, which forms the causal chain. Each evidence $E_i$ consists of a start time $t_{start, i}$, an end time $t_{end, i}$, a textual rationale $R_i$, and a variable-length map of bounding boxes $B_i$. This map $B_i = \{s_j : b_j\}_{j=1}^{P_i}$ contains $P_i$ entries, where each $s_j$ is a 1-FPS timestamp within the evidence's time range $[t_{start, i}, t_{end, i}]$, and $b_j = [x_{min}, y_{min}, x_{max}, y_{max}]$ is the corresponding spatial bounding box. Detailed prompt for VLMs is provided in supplementary materials (Sec. \hyperref[prompt7:cast_bench_evaluation_prompt]{10.1}).

\section{CaST-Bench}
\label{sec:cast_bench}

To evaluate this \underline{Ca}usal Chain-Grounded \underline{S}patio-\underline{T}emporal Reasoning task, we present CaST-Bench. Creating such a benchmark is challenging due to the need to jointly generate causal chains in QA pairs and fine-grained visual evidence, including explicit temporal segments and spatial bounding boxes. To address this, we propose a multi-stage, human-AI collaborative pipeline for dataset construction in \cref{sec:construction} and \cref{sec:stat}. Then, we introduce a novel evaluation suite to rigorously test model's capabilities in \cref{sec:eval}.

\subsection{Dataset Construction}
\label{sec:construction}
\begin{figure*}
    \centering
    \includegraphics[width=0.99\linewidth]{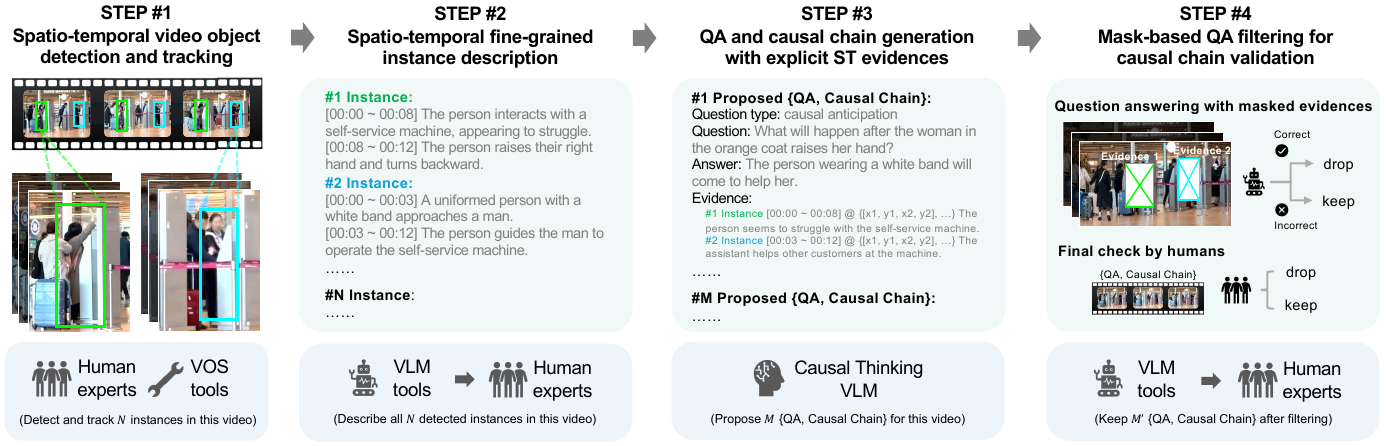}
    \caption{\textbf{Human-AI Collaborative Pipeline for Constructing Causal Chain–Grounded Video QA.} Vision-language tools and humans collaborate across detection, description, generation, and filtering stages to produce high-quality \{QA, Causal Chain\} data.}
    \label{fig:generation_pipeline}
    \vspace{-1em}
\end{figure*}

Our dataset construction follows a multi-stage pipeline, as shown in \cref{fig:generation_pipeline}, where VLM refers to Gemini-2.5-pro \cite{comanici2025gemini}.

\paragraph{Video Selection}
To ensure reasoning difficulty, we avoid curated videos, such as movies and TV shows, whose frame composition simplifies spatial grounding. Instead, we select naturalistic videos from the SegmentAnything-Video (SAV) dataset \cite{ravi2024sam2}, which provides expert-annotated dense instance segmentations and tracking. We then use a VLM to filter and retain videos containing multiple objects engaged in distinct activities, ensuring complex scenes that pose a challenge for spatio-temporal reasoning. This process yields 1,015 qualified videos.

\paragraph{Spatio-Temporal Fine-grained Instance Description}
To build a rich corpus for subsequent causal QA generation, we produce fine-grained spatio-temporal descriptions for each tracked instance (see the second part of \cref{fig:generation_pipeline}). A key challenge is that our videos contain numerous small objects, and existing VLMs struggle to describe them accurately due to interference from surrounding objects. To address this, we design a pipeline in which a VLM first generates descriptions, and human experts then validate and revise them.
Specifically, we craft sophisticated visual and textual prompts that combine static full frames with cropped videos around the target instances, enabling the VLM to generate more accurate action descriptions (details in the supplementary material \cref{sec:supp_pipeline_spatio_temporal}). Human experts then review and rewrite these AI-generated captions using a custom annotation tool. This process yields 10,728 high-quality dynamic instance descriptions.

\begin{figure*}[t]
  \centering
  \begin{minipage}[t]{0.33\linewidth}
    \vspace{0pt} 
    \centering
    \includegraphics[width=\linewidth]{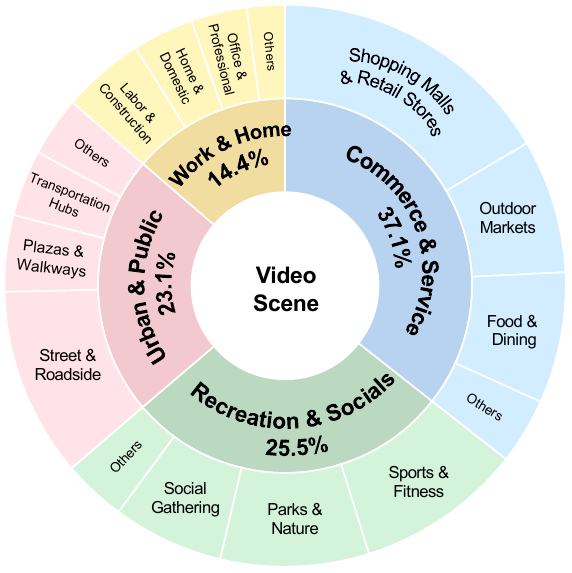}
    \captionof{figure}{Distribution of Video Categories.}
    \label{fig:video-dist}
  \end{minipage}
  \hfill
  \begin{minipage}[t]{0.33\linewidth}
    \vspace{0pt}
    \centering
    \includegraphics[width=\linewidth]{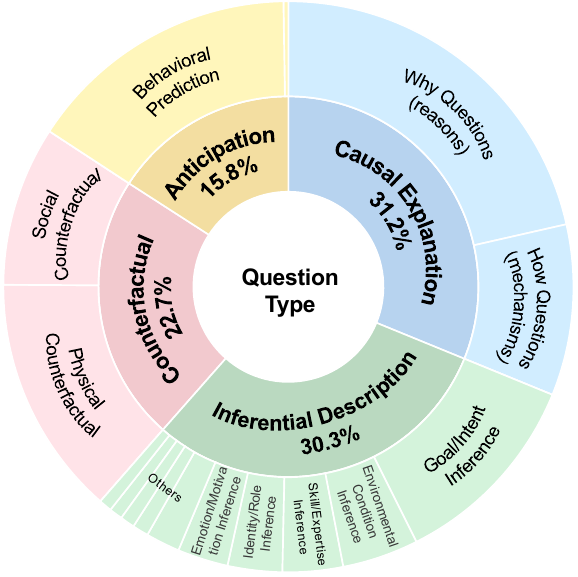}
    \captionof{figure}{Distribution of Question Types.}
    \label{fig:question-types}
  \end{minipage}
  \hfill
  \begin{minipage}[t]{0.3\linewidth}
  \vspace{30pt}
  \centering
  \scalebox{0.8}{
    \rowcolors{2}{lightblue!30}{white}
    \begin{tabular}{l r}
      \hline
      \rowcolor{lightblue!60}
      \multicolumn{2}{c}{\textbf{Benchmark Statistics}} \\
      \#Videos & 1015 \\
      \#Described instances & 10728 \\
      \#Questions & 2066 \\
      \#Options per QA & 6 \\
      Avg. \#evidences per QA & 2.36 \\
      Avg. video tem. duration & 13.68s \\
      Avg. evidence tem. duration & 5.65s \\
      Avg. evidence spa. coverage & 8.3\% \\
      Avg. \#words of questions & 21.03 \\
      Avg. \#words of options & 10.86 \\
      Avg. \#words of evidence & 15.50 \\
      \hline
    \end{tabular}
  }
  \vspace{18pt}
  \captionof{table}{Detailed Benchmark Statistics.}
  \label{tab:annotation-stats}
\end{minipage}

\vspace{-1em}
\end{figure*}

\paragraph{QA and Causal Chain Generation with Explicit ST Evidences}
Leveraging the dense instance descriptions, we prompt a state-of-the-art VLM, as a causal thinking agent, to identify potential causal relationships and generate structured QA pairs with causal chains. To ensure diversity, we design a comprehensive taxonomy of causal reasoning questions categorized into four main types (the specific subcategories for each main type are detailed in \cref{fig:question-types} and the supplementary material \cref{sec:supp_pipeline_qa_and_causal}):
\begin{itemize}
    \item \textbf{Causal Explanation} Questions that explain the reasons (\textit{why}) or mechanisms (\textit{how}) behind actions or events.
    \item \textbf{Counterfactual Reasoning} Questions that infer the consequence of a single alteration to a key causal element.
    \item \textbf{Predictive Anticipation} Questions that require predicting the most probable and immediate outcome.
    \item \textbf{Inferential Description} Questions that infer implicit attributes or states (e.g., roles, intentions, emotions).
\end{itemize}
In summary, the causal thinking VLM takes as input the instance descriptions, the video, and the question-type taxonomy, and outputs structured question-answer pairs along with their associated causal chains.

\paragraph{Multiple-Choice Question Generation}
To construct multiple-choice questions from the generated QA pairs, we introduce distractor options designed to challenge models beyond surface-level understanding. As shown in \cref{fig:causal_graph}, mitigating the effect of confounders is fundamental to causal inference. Therefore, we generate two types of distractors based on different confounding sources: 1)	Text-based distractors: We prompt a LLM to generate plausible yet incorrect answers using only the question text, without access to the video content. 2)	Video-based distractors: We identify causally irrelevant visual instances present in the video to create distractors that exploit visual biases. If a model selects any of these distractors, it indicates reliance on spurious linguistic or visual cues rather than genuine visual evidence. This design not only increases the benchmark’s difficulty but also enables fine-grained error analysis.

\paragraph{Mask-Based QA Filtering for Causal Chain Validation}
We adopt a three-step filtering procedure to ensure both question difficulty and causal-chain completeness. First, we apply a text-based filter to remove QA pairs that an LLM can answer correctly without the video. Second, we introduce a novel video-based filter that masks all spatio-temporal regions containing the causal evidences. A VLM is then prompted to answer the question using the masked video. If it still succeeds, the result indicates that the causal chain is incomplete or the question can be solved through dataset biases or other unintended cues; such items are discarded. Finally, human annotators conduct a thorough review to validate the question, answer, and causal chain. After these rounds of filtering, only 40\% of the initial QA pairs remain.

\subsection{Dataset Statistics}
\label{sec:stat}

1,015 videos are selected from the SegmentAnything-Video \cite{ravi2024sam2} dataset, covering a broad range of countries, cultures, and environments. To categorize scene types, we use a VLM to label videos, confirming that the benchmark spans diverse real-world scenarios (see \cref{fig:video-dist}). From this video set, we generated 2,066 causal questions across all taxonomy-defined question types (see \cref{fig:question-types}). Each question is paired with an answer and a fully grounded causal chain that links linguistic reasoning to the corresponding spatio-temporal visual evidence. Detailed statistics are presented in \cref{tab:annotation-stats}.

\subsection{Evaluation Suite}
\label{sec:eval}

\vspace{-0.2em}

Since our task of Causal Chain-Grounded Spatio-Temporal Reasoning is novel, we design a new evaluation suite to assess the accuracy of both the grounded causal chain and the final answer. Unlike previous single-target spatio-temporal grounding tasks \cite{cheng2025vstarbenchmark, zhang2020videostg}, we evaluate the grounded causal chain that may involve multiple temporal segments and spatial boxes for different instances. Our metrics address this challenge by first matching predicted and ground-truth instances before computing a series of metrics, as follows.

\paragraph{Grounded Causal Chain Evaluation}
To quantitatively evaluate the alignment between a predicted causal chain and the ground-truth causal chain, we first use a greedy matching algorithm to pair predicted instances with GT instances based on their spatio-temporal overlap (see supplementary material \cref{sec:supp_evaluation_suite} for details). For the matched pairs, we propose two metrics: Instance-Matched tIoU (IM-tIoU) for temporal accuracy and Instance-Matched vIoU (IM-vIoU) for spatio-temporal accuracy.

The IM-tIoU is defined as:
\begin{equation}
\label{eq:m_tiou}
\scalebox{0.8}{$\displaystyle \text{IM-tIoU} = \frac{1}{|G|} \sum_{(p,g) \in M} \frac{|\text{span}(p) \cap \text{span}(g)|}{|\text{span}(p) \cup \text{span}(g)|}$},
\end{equation}
where $G$ is the set of GT instances, $M$ is the set of matched prediction ($p$) and GT ($g$) instance pairs, and $\text{span}(\cdot)$ denotes the temporal intervals of all evidences about an instance.

Our IM-vIoU metric is defined as:
\begin{equation}
\label{eq:m_viou}
\scalebox{0.8}{$\displaystyle \text{IM-vIoU} = \frac{1}{|G|} \sum_{(p,g) \in M} \left( \frac{|\text{span}(p) \cap \text{span}(g)|}{|\text{span}(p) \cup \text{span}(g)|} \cdot \frac{1}{|\mathcal{T}_{ov}|} \sum_{t \in \mathcal{T}_{ov}} \text{sIoU}(p_t, g_t) \right)$},
\end{equation}
where $\mathcal{T}_{ov}$ is the set of overlapping frames between predicted instance $p$ and GT instance $g$, and $\text{sIoU}(p_t, g_t)$ is the spatial IoU of their bounding boxes at frame $t$.

\paragraph{Comprehensive Answer Evaluation}
To excel on CaST-Bench, a model must not only correctly ground the causal chain but also produce the correct final answer. We introduce the Faithful Answer Rate ($\mathcal{F}$) and Spurious Answer Rate ($\mathcal{S}$) to jointly evaluate causal chain grounding and answer correctness. $\mathcal{F}$ measures how often a correct answer is supported by well-grounded evidence, whereas $\mathcal{S}$ measures how often a correct answer is given despite poor grounding.
\begin{equation}
\label{eq:faithful_spurious}
\scalebox{0.8}{$\displaystyle
\begin{aligned}
\mathcal{F} &= \mathbb{E}[\mathbb{I}(\text{Acc}=1) \cdot \mathbb{I}(\text{IM-vIoU} \ge \tau_{st})], \\
\mathcal{S} &= \mathbb{E}[\mathbb{I}(\text{Acc}=1) \cdot \mathbb{I}(\forall p \in P, \text{vIoU}_p < \tau_{st})],
\end{aligned}
$}
\end{equation}
where $\mathbb{I}(\cdot)$ is the indicator function, Acc is the answer accuracy, and $\tau_{st}$ is a predefined threshold.

\paragraph{Open-Ended Evaluation} To enable analysis along more dimensions, we adopt an LLM-as-a-judge framework to compare both the final answer and the rationales in the causal chain ($E_c$) with the ground-truth. We design four criteria: \textit{answer correctness}, which evaluates the semantic correctness of the open-ended final answer; \textit{logical consistency}, which assesses the coherence between the causal chain and the final answer; \textit{evidence coverage}, which measures whether all ground-truth evidences are covered; and \textit{overall justification}, which provides a holistic assessment of both the causal chain and the final answer. Detailed definitions are provided in the supplementary material \cref{sec:supp_evaluation_suite_open_ended}.
\section{Experiments}

\begin{table*}[t]
\centering
\renewcommand{\arraystretch}{1.22}
\setlength{\tabcolsep}{4pt}
\small
{%
\resizebox{\textwidth}{!}{%
\begin{tabular}{@{}l*8{c}@{}}
\toprule
\multicolumn{1}{l}{\multirow{3}{*}{\textbf{Model}}} &
\multirow{2}{*}{\rule{0pt}{2.4ex}\textbf{Answer Accuracy (MCQ)}} &
\multicolumn{4}{c}{\textbf{Evidence Grounding}} &
\multicolumn{2}{c}{\textbf{Evidence--Answer Faithfulness}} &
\multirow{2}{*}{\rule{0pt}{2.4ex}\textbf{Open-Ended Evaluation}} \\
\cmidrule(lr){3-6}\cmidrule(lr){7-8}
& 
  & \multicolumn{2}{c}{\textbf{Temporal}} &
  \multicolumn{2}{c}{\textbf{Spatio-Temporal}} &
  \multirow{2}{*}{\rule{0pt}{2.4ex}\textbf{Faithful Rate}} &
  \multirow{2}{*}{\rule{0pt}{2.4ex}\textbf{Spurious Rate}} &
  \\ 
\cmidrule(lr){3-4}\cmidrule(lr){5-6}
& & \textbf{R@0.5} & \textbf{IM-tIoU} & \textbf{R@0.1} & \textbf{IM-vIoU} & & & \\
\midrule
\rowcolor[gray]{0.96}\multicolumn{9}{l}{\textbf{\textit{Baselines}}}\\
Random
& 16.67 & -- & -- & -- & -- & -- & -- & -- \\
Gemini-2.5-Pro (text-only)
& 23.14 & -- & -- & -- & -- & -- & -- & -- \\
Human
& 91.89 & -- & -- & -- & -- & -- & -- & -- \\
\midrule
\rowcolor[gray]{0.96}\multicolumn{9}{l}{\textbf{\textit{Proprietary Models}}}\\
Gemini-2.5-Pro \cite{comanici2025gemini}
& \textbf{50.34} & 23.29 & 21.53 & 8.14 & 2.46 & 7.60 & 42.26 & 3.85 \\
Gemini-2.5-Flash \cite{comanici2025gemini}
& 45.60 & \textbf{29.45} & \textbf{27.63} & 13.29 & 3.52 & 9.97 & 33.35 & 3.83 \\
GPT-5 \cite{openai_gpt5_2025}
& 46.32 & 27.83 & 26.61 & \textbf{16.19} & \textbf{4.31} & \textbf{12.68} & 32.91 & \textbf{4.30} \\
GPT-5 mini \cite{openai_gpt5_2025}
& 37.22 & 21.28 & 19.89 & 8.31 & 2.41 & 6.29 & \textbf{30.83} & 3.98 \\
\midrule
\rowcolor[gray]{0.96}\multicolumn{9}{l}{\textbf{\textit{Open-Source Models}}}\\
GLM-4.1V-9B-Thinking \cite{hong2025glm}
& 39.55 & 17.50 & 16.58 & 6.18 & 1.93 & 4.74 & 34.32 & 2.66 \\
InternVL-3.5-30B-A3B \cite{wang2025internvl3_5}
& 44.53 & 9.04 & 8.33 & 0.42 & 0.25 & 0.48 & 44.00 & 2.49 \\
InternVL-3.5-14B \cite{wang2025internvl3_5}
& 43.27 & 9.24 & 8.77 & 0.56 & 0.33 & 0.73 & 42.55 & 2.50 \\
InternVL-3.5-8B \cite{wang2025internvl3_5}
& 40.95 & 9.40 & 8.59 & 0.61 & 0.28 & 0.58 & 40.37 & 2.52 \\
InternVL-2.5-8B \cite{chen2024internvl25}
& 39.55 & -- & -- & -- & -- & -- & -- & 2.59 \\
Qwen3-VL-8B-Thinking \cite{qwen3vl2025} & 39.16 & 10.44 & 10.23 & 2.69 & 0.87 & 2.27 & 36.74 & 2.15 \\
Qwen3-VL-8B-Instruct \cite{qwen3vl2025} & 43.13 & 10.63 & 10.51 & 3.41 & 1.07 & 2.76 & 39.84 & 1.93 \\
Qwen3-VL-4B-Instruct \cite{qwen3vl2025} & 45.30 & 11.65 & 11.21 & 2.94 & 0.93 & 2.76 & 42.30 & 2.56 \\ 
Qwen2.5-VL-7B-Instruct \cite{Qwen2.5-VL}
& 41.09 & 3.80 & 3.72 & 0.13 & 0.09 & 0.29 & 40.80 & 1.94 \\
MiMo-VL-7B-RL-2508 \cite{coreteam2025mimovltechnicalreport}
& 32.24 & 10.42 & 9.63 & 2.74 & 0.81 & 2.81 & 31.43 & 2.84 \\
LLaVA-NeXT-Video-34B \cite{zhang2024llavanextvideo}
& 28.17 & -- & -- & -- & -- & -- & -- & 1.39 \\
\bottomrule
\end{tabular}%
}
}
\caption{
\textbf{Comprehensive comparison of proprietary and open-source vision-language models on \textit{CaST-Bench}, our benchmark designed to evaluate causal chain-grounded spatio-temporal reasoning.}  
The evaluation spans four dimensions: 
(1) \textbf{Answer Accuracy} multiple-choice question (MCQ) score; 
(2) \textbf{Evidence Grounding} assessing how accurately models construct and utilize spatio-temporal evidence chains; 
(3) \textbf{Evidence--Answer Faithfulness} measuring the causal alignment between generated evidences and final answers; 
and (4) \textbf{Open-Ended Evaluation} capturing overall justification quality.
“--” indicates not applicable or failure to produce valid output. 
}
\label{tab:main_table}
\end{table*}

\vspace{-1mm}
\subsection{Settings}
\vspace{-1mm}

\paragraph{Models} We evaluate 4 proprietary models (Gemini-2.5-Pro, Gemini-2.5-Flash \cite{comanici2025gemini} and GPT-5, GPT-5 mini \cite{openai_gpt5_2025}) and variants of 7 open-source VLMs (GLM \cite{hong2025glm}, MiMo \cite{coreteam2025mimovltechnicalreport},  InternVL-3.5 \cite{wang2025internvl3_5},  InternVL-2.5 \cite{chen2024internvl25}, Qwen3-VL \cite{qwen3vl2025}, Qwen2.5-VL \cite{Qwen2.5-VL}, LLaVA-NeXT-Video \cite{zhang2024llavanextvideo}).

\vspace{-3mm}

\paragraph{Implementation Details} Proprietary models are evaluated via official APIs, while open-source models are run locally with LLaMA-Factory \cite{zheng2024llamafactory}. 
Models with native video support take raw videos as input; others (i.e., the GPT-5 series) receive uniformly sampled 1-FPS frames. 
All models must generate a grounded ST causal chain before producing the final answer. When computing metrics, we set $\tau_{st}$ to 0.1, and use \cite{comanici2025gemini} as the LLM judge for open-ended evaluation. Additional details are provided in the supplementary material \cref{sec:supp_evaluation_suite}.

\vspace{-1mm}

\subsection{Main Results}

\vspace{-1.8mm}

\paragraph{Overall Results} Across models, the Answer Accuracy scores in Table~\ref{tab:main_table} remain substantially below human baseline.
Because \textit{CaST-Bench} is deliberately constructed around questions that require accurate ST grounding, these low scores indicate that current models struggle to identify and interpret the relevant visual evidence in the video.
At the same time, generating explicit ST evidence poses an additional challenge, as it requires not only identifying the correct visual cues but also expressing them as ST evidences.
This additional step may partly account for the low grounding scores.


\begin{table*}[t]
\raggedright
\renewcommand{\arraystretch}{1.22}
\setlength{\tabcolsep}{4pt}
\small

\begin{minipage}[t]{0.48\textwidth}
\centering
\resizebox{0.95\textwidth}{!}{
\begin{tabular}{@{}lccc@{}}
\toprule
\textbf{Model} &
\shortstack{\textbf{w/o}\\\textbf{Causal Chain}} &
\shortstack{\textbf{w/}\\\textbf{Causal Chain}} &
\shortstack{\textbf{Given GT}\\\textbf{Causal Chain}} \\
\midrule
\rowcolor[gray]{0.96}\multicolumn{4}{@{}l@{}}{\textbf{\textit{Proprietary Models}}}\\
Gemini-2.5-Pro & 45.98 & 50.34 & 75.61 \\
GPT-5          & 43.90 & 46.32 & 71.88 \\
\midrule
\rowcolor[gray]{0.96}\multicolumn{4}{@{}l@{}}{\textbf{\textit{Open-Source Models}}}\\
GLM-4.1V-9B & 38.09 & 39.55 & 65.25 \\
Qwen2.5-VL-7B & 40.90 & 41.09 & 59.68 \\
\bottomrule
\end{tabular}}
\vspace{2mm}
\caption{
\textbf{Ablation study on the effect of causal chain evidences on answer accuracy (\%).}  
Results on \textit{CaST-Bench} under three setups:  
(1) \textbf{w/o Causal Chain}: direct answering;  
(2) \textbf{w/ Causal Chain}: reasoning via self-generated chains;  
(3) \textbf{Given GT Causal Chain}: using ground-truth chains.
}
\label{tab:ablation_gt}
\end{minipage}
\hfill
\begin{minipage}[t]{0.48\textwidth}
\centering
\resizebox{\textwidth}{!}{
\begin{tabular}{@{}l
>{\centering\arraybackslash}m{0.17\textwidth}
>{\centering\arraybackslash}m{0.17\textwidth}
>{\centering\arraybackslash}m{0.17\textwidth}
>{\centering\arraybackslash}m{0.17\textwidth}@{}}
\toprule
\textbf{Model} &
\shortstack{\textbf{CE}\\\vphantom{Given GT Causal Chain}} &
\shortstack{\textbf{CR}\\\vphantom{Given GT Causal Chain}} &
\shortstack{\textbf{PA}\\\vphantom{Given GT Causal Chain}} &
\shortstack{\textbf{ID}\\\vphantom{Given GT Causal Chain}} \\
\midrule
\rowcolor[gray]{0.96}\multicolumn{5}{@{}l@{}}{\textbf{\textit{Proprietary Models}}}\\
Gemini-2.5-Pro & \textbf{56.83} & \textbf{34.54} & \textbf{51.38} & \textbf{54.95} \\
GPT-5          & 50.00 & 32.62 & 49.24 & 51.28 \\
\midrule
\rowcolor[gray]{0.96}\multicolumn{5}{@{}l@{}}{\textbf{\textit{Open-Source Models}}}\\
GLM-4.1V-9B      & 39.44 & 29.42 & 42.81 & 45.53 \\
Qwen2.5-VL-7B    & 42.39 & 31.34 & 44.34 & 45.37 \\
\bottomrule
\end{tabular}}
\vspace{2mm}
\caption{
\textbf{Performance across question categories on \textit{CaST-Bench}.}  
Columns report multiple-choice answer accuracy (\%) for four reasoning categories:  
\textbf{CE} = \textit{Causal Explanation},  
\textbf{CR} = \textit{Counterfactual Reasoning},  
\textbf{PA} = \textit{Predictive Anticipation},  
\textbf{ID} = \textit{Inferential Description}.
}
\label{tab:ablation_question_categories}
\end{minipage}
\vspace{-0.5em}
\end{table*}

\vspace{-2mm}
\paragraph{Baselines and the Necessity of Visual Evidence}
As shown in Table~\ref{tab:main_table}, random guessing yields 16.67\% and text-only input for Gemini-2.5-Pro reaches only 23.14\%, while human annotators achieve 91.89\%. 
This large performance gap highlights the difficulty of \textit{CaST-Bench} and the necessity of visual evidence for causal question answering.


\vspace{-2mm}
\paragraph{Comparison Between Proprietary and Open-Source Models}
Across all evaluation dimensions, proprietary models consistently outperform open-source ones. 
Gemini-2.5-Pro, Gemini-2.5-Flash, and GPT-5 achieve higher MCQ accuracy (up to 50.34\%) and notably stronger evidence grounding and faithful answer rates, whereas open-source models generally remain in the 30–45\% MCQ accuracy range with much weaker grounding. 
These results indicate that, despite recent progress, open-source systems still fall substantially short on deeper causal reasoning that requires grounded spatio-temporal understanding.

\vspace{-2mm}
\paragraph{Evidence Grounding and Faithfulness}
Despite moderate MCQ accuracy, all models struggle to ground the causal chain. 
Proprietary models reach only 20--28\% in IM-tIoU and 2--4\% in IM-vIoU, while open-source models perform even worse, indicating that precise spatio-temporal grounding is largely absent. 
As a result, faithful answer rates remain low (e.g., 12.68\% for GPT-5 and mostly below 5\% for open-source models), whereas spurious rates stay high (30--45\%), showing that many correct answers arise from poorly grounded or irrelevant evidence chains.

\vspace{-2mm}
\paragraph{Model-Level Observations}
We outline several observations drawn from the evaluated models.
First, ``thinking'' variants do not consistently outperform their ``instruct'' counterparts. For example, Qwen3-VL-8B-Instruct attains higher answer accuracy than Qwen3-VL-8B-Thinking, suggesting that explicit textual ``thinking'' traces do not necessarily yield stronger causal-chain reasoning. This may stem from repetitive thinking content that leads thinking models to omit final answers or from their language-centric reasoning disrupting the visual–linguistic alignment required by CaST-Bench. More failure details can be found in the supplementary material \cref{sec:supp_failure}.

Second, model scaling yields mixed effects: InternVL-3.5-30B-A3B improves MCQ accuracy over its smaller variants but does not exhibit corresponding gains in grounding or faithfulness, indicating that increased capacity alone does not guarantee better causal-chain understanding.   

Third, within several model families (Gemini, GPT, InternVL), larger variants exhibit higher spurious rates. This may indicate that, on the one hand, larger models are better able to choose a likely answer based on the given information; on the other hand, they may indeed attend to the relevant spatio-temporal cues but lack the ability to express them faithfully, both of which can contribute to higher spurious rates.

Finally, some models (e.g., InternVL-2.5-8B, LLaVA-NeXT-Video-34B) frequently fail to follow the required structured-output format, resulting in skipped examples and missing grounding scores (``--'').

\begin{figure*}[t]
  \centering
  \begin{minipage}[t]{0.473\linewidth}
    \vspace{0pt}
    \centering
    \includegraphics[width=\linewidth]{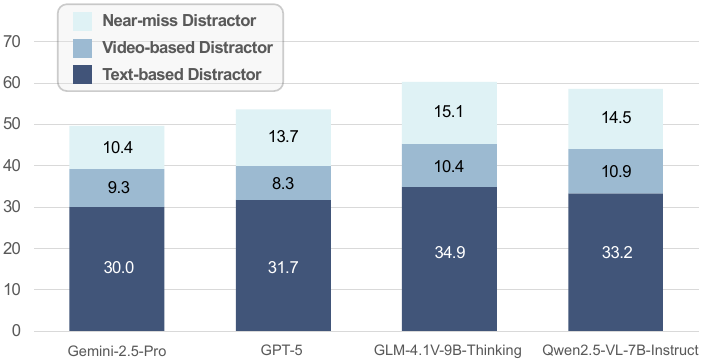}
    \captionof{figure}{\textbf{Error analysis regarding model vulnerability to different distractor option types on \textit{CaST-Bench}.}
    Values show how often model selected the distractor option (i.e., the trap rate)
    for text-based, video-based, and near-miss distractors.}
    \label{fig:error-analysis-distractor}
  \end{minipage}
  \hfill
  \begin{minipage}[t]{0.473\linewidth}
    \vspace{0pt}
    \centering
    \includegraphics[width=\linewidth]{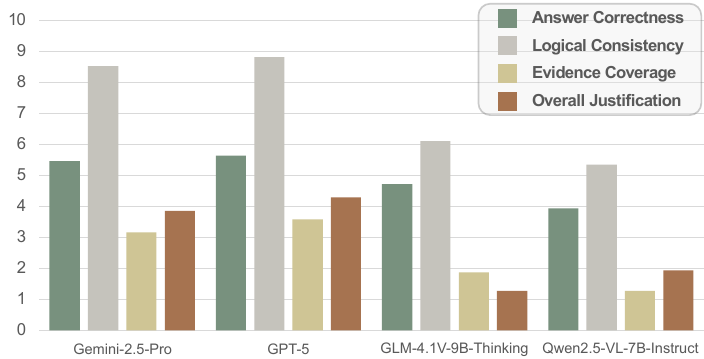}
    \captionof{figure}{\textbf{Open-ended evaluation result on \textit{CaST-Bench}.} We evaluate each model along three open-ended dimensions, \textbf{Answer Correctness}, \textbf{Logical Consistency}, and \textbf{Evidence Coverage}, as well as the overall dimension of \textbf{Overall Justification}.}
    \label{fig:error-analysis-open-ended}
  \end{minipage}
  \vspace{-1em}
\end{figure*}

\subsection{Ablation \& Analysis}

\paragraph{Effect of Causal Chain Evidences}
Table~\ref{tab:ablation_gt} presents an ablation study on how causal chain evidences affect answer accuracy on \textit{CaST-Bench}. 
Across all models, incorporating causal-chain reasoning (i.e., \textbf{w/ Causal Chain}) consistently improves performance compared with direct answering (i.e., \textbf{w/o Causal Chain}), indicating that explicit step-by-step causal reasoning in spatio-temporal space helps models derive more accurate answers.
Providing ground-truth causal chains (i.e., \textbf{Given GT Causal Chain}) yields a substantial further improvement—about 20--25 points across both proprietary and open-source models. Importantly, we use an LLM to remove any statements from GT evidence rationale that could reveal or suggest the final answer and retain only the factual descriptions of the spatio-temporal evidence, ensuring that the gain reflects the contribution of correct causal grounding rather than answer leakage. This 20--25 points improvement reveals a large and consistent gap between the models’ self-generated causal chain and the ideal causal chain for reasoning, suggesting that current models, regardless of scale or origin, still struggle to produce reliable causal chains and that causal-chain understanding remains a primary bottleneck in multimodal causal reasoning.

\vspace{-2mm}
\paragraph{Performance across Question Categories}
Table~\ref{tab:ablation_question_categories} reports model performance across four reasoning categories on \textit{CaST-Bench}. 
Proprietary models (Gemini-2.5-Pro and GPT-5) consistently outperform open-source ones across all categories, suggesting stronger multimodal alignment and causal reasoning capabilities. 
Among the four types, \textbf{Causal Explanation (CE)} questions yield the highest accuracies, indicating that models are relatively good at capturing explicit causal cues from visual evidence. 
In contrast, \textbf{Counterfactual Reasoning (CR)} remains the most challenging category, with accuracies around 30\%, as it requires hypothetical interventions and deeper causal inference beyond direct observation. 
\textbf{Predictive Anticipation (PA)} and \textbf{Inferential Description (ID)} lie in between, reflecting that models can extrapolate temporal trends and infer underlying relations to some extent, yet still lack fully grounded spatio-temporal reasoning. 
Overall, the results suggest that robustness of causal reasoning remains an open challenge for multimodal systems.

\subsection{Error Analysis}

\paragraph{Distractor-Level Error Patterns} \cref{fig:error-analysis-distractor} shows that across all models, text-based distractors are the most misleading, yielding the highest trap rates and indicating that models often favor plausible textual alternatives over the actual visual evidence.
In contrast, video-based distractors exhibit the lowest trap rates. Near-miss distractors fall in between, highlighting persistent challenges in fine-grained causal reasoning.
Finally, proprietary models (e.g., GPT-5, Gemini-2.5-Pro) are consistently more robust, while open-source models show higher vulnerability across all distractor types.

\vspace{-2mm}
\paragraph{Open-Ended Error Patterns} As shown in \cref{fig:error-analysis-open-ended}, proprietary models outperform open-source models across all evaluated dimensions, indicating consistently stronger performance across answer correctness, logical consistency, evidence coverage, and overall justification. Logical consistency receives the highest scores for all models, reflecting their general ability to produce coherent natural-language explanations. In contrast, evidence coverage is the weakest dimension, revealing persistent difficulty in identifying and referencing the visual cues required for causal reasoning. A similar pattern is observed in the MCQ setting.  Open-source models show lower scores in evidence-grounded reasoning, which is also reflected in their lower overall justification performance.

\section{Conclusion}

We introduce CaST-Bench, a new benchmark for Causal Chain-Grounded Spatio-Temporal Video QA, featuring 2,066 complex questions annotated with fine-grained spatio-temporal causal chains. Accompanying the benchmark, we design a comprehensive evaluation suite to assess answer correctness, evidence grounding, and evidence-answer faithfulness. This benchmark addresses a critical gap, as prior works often overlooked the chained spatio-temporal grounding essential for genuine causal reasoning. Our experiments demonstrate that current VLMs struggle significantly on this task, largely due to their inability to construct precise causal chains. This failure to explicitly locate and align causal evidence highlights the unique diagnostic value of CaST-Bench. We further confirmed that incorporating accurately grounded causal chains is a crucial step that substantially improves answer accuracy. We hope CaST-Bench will support the development of future VLMs capable of deeper, more transparent, and causally grounded video understanding.




\section{Acknowledgement}
{
This work was supported by JPNP20017, subsidized by the New Energy and Industrial Technology Development Organization (NEDO), JSPS KAKENHI Grant Number \\ JP24K02956, 25K24384,
JST ASPIRE Grant Number \\ JPMJAP2303.
}
{
    \small
    \bibliographystyle{ieeenat_fullname}
    \bibliography{main}

\begin{thebibliography}{51}
\providecommand{\natexlab}[1]{#1}
\providecommand{\url}[1]{\texttt{#1}}
\expandafter\ifx\csname urlstyle\endcsname\relax
  \providecommand{\doi}[1]{doi: #1}\else
  \providecommand{\doi}{doi: \begingroup \urlstyle{rm}\Url}\fi

\bibitem[Bai et~al.(2025)Bai, Chen, Liu, Wang, Ge, Song, Dang, Wang, Wang, Tang, Zhong, Zhu, Yang, Li, Wan, Wang, Ding, Fu, Xu, Ye, Zhang, Xie, Cheng, Zhang, Yang, Xu, and Lin]{Qwen2.5-VL}
Shuai Bai, Keqin Chen, Xuejing Liu, Jialin Wang, Wenbin Ge, Sibo Song, Kai Dang, Peng Wang, Shijie Wang, Jun Tang, Humen Zhong, Yuanzhi Zhu, Mingkun Yang, Zhaohai Li, Jianqiang Wan, Pengfei Wang, Wei Ding, Zheren Fu, Yiheng Xu, Jiabo Ye, Xi Zhang, Tianbao Xie, Zesen Cheng, Hang Zhang, Zhibo Yang, Haiyang Xu, and Junyang Lin.
\newblock Qwen2.5-vl technical report.
\newblock \emph{arXiv preprint arXiv:2502.13923}, 2025.

\bibitem[Chen et~al.(2024{\natexlab{a}})Chen, Liu, Huang, He, Pei, Xu, Wang, Lu, and Wang]{chen2024cgbench}
Guo Chen, Yicheng Liu, Yifei Huang, Yuping He, Baoqi Pei, Jilan Xu, Yali Wang, Tong Lu, and Limin Wang.
\newblock Cg-bench: Clue-grounded question answering benchmark for long video understanding.
\newblock \emph{arXiv preprint arXiv:2412.12075}, 2024{\natexlab{a}}.

\bibitem[Chen et~al.(2024{\natexlab{b}})Chen, Liu, He, Chen, Gan, Ma, Zhong, Zhang, Wang, Lin, et~al.]{chen2024mecd}
Tieyuan Chen, Huabin Liu, Tianyao He, Yihang Chen, Chaofan Gan, Xiao Ma, Cheng Zhong, Yang Zhang, Yingxue Wang, Hui Lin, et~al.
\newblock Mecd: Unlocking multi-event causal discovery in video reasoning.
\newblock \emph{Advances in Neural Information Processing Systems}, 37:\penalty0 92554--92580, 2024{\natexlab{b}}.

\bibitem[Chen et~al.(2025{\natexlab{a}})Chen, Liu, Chen, Su, Zheng, and Lin]{chen2025cross}
Weixing Chen, Yang Liu, Binglin Chen, Jiandong Su, Yongsen Zheng, and Liang Lin.
\newblock Cross-modal causal relation alignment for video question grounding.
\newblock In \emph{Proceedings of the Computer Vision and Pattern Recognition Conference}, pages 24087--24096, 2025{\natexlab{a}}.

\bibitem[Chen et~al.(2025{\natexlab{b}})Chen, Ge, Wang, Ge, Qiu, Shan, and Liu]{chen2025exploring}
Yi Chen, Yuying Ge, Rui Wang, Yixiao Ge, Lu Qiu, Ying Shan, and Xihui Liu.
\newblock Exploring the effect of reinforcement learning on video understanding: Insights from seed-bench-r1.
\newblock \emph{arXiv preprint arXiv:2503.24376}, 2025{\natexlab{b}}.

\bibitem[Chen et~al.(2024{\natexlab{c}})Chen, Wang, Cao, Liu, Gao, Cui, Zhu, Ye, Tian, Liu, et~al.]{chen2024internvl25}
Zhe Chen, Weiyun Wang, Yue Cao, Yangzhou Liu, Zhangwei Gao, Erfei Cui, Jinguo Zhu, Shenglong Ye, Hao Tian, Zhaoyang Liu, et~al.
\newblock Expanding performance boundaries of open-source multimodal models with model, data, and test-time scaling.
\newblock \emph{arXiv preprint arXiv:2412.05271}, 2024{\natexlab{c}}.

\bibitem[Cheng et~al.(2025{\natexlab{a}})Cheng, Ge, Wang, Ge, Liao, and Shan]{cheng2025videoholmes}
Junhao Cheng, Yuying Ge, Teng Wang, Yixiao Ge, Jing Liao, and Ying Shan.
\newblock Video-holmes: Can mllm think like holmes for complex video reasoning?
\newblock \emph{arXiv preprint arXiv:2505.21374}, 2025{\natexlab{a}}.

\bibitem[Cheng et~al.(2025{\natexlab{b}})Cheng, Hu, Liu, Si, Li, and Gong]{cheng2025vstarbenchmark}
Zixu Cheng, Jian Hu, Ziquan Liu, Chenyang Si, Wei Li, and Shaogang Gong.
\newblock V-star: Benchmarking video-llms on video spatio-temporal reasoning, 2025{\natexlab{b}}.

\bibitem[Comanici et~al.(2025)Comanici, Bieber, Schaekermann, Pasupat, Sachdeva, Dhillon, Blistein, Ram, Zhang, Rosen, et~al.]{comanici2025gemini}
Gheorghe Comanici, Eric Bieber, Mike Schaekermann, Ice Pasupat, Noveen Sachdeva, Inderjit Dhillon, Marcel Blistein, Ori Ram, Dan Zhang, Evan Rosen, et~al.
\newblock Gemini 2.5: Pushing the frontier with advanced reasoning, multimodality, long context, and next generation agentic capabilities.
\newblock \emph{arXiv preprint arXiv:2507.06261}, 2025.

\bibitem[Feng et~al.(2025)Feng, Gong, Li, Guo, Wang, Peng, Wu, Zhang, Wang, and Yue]{feng2025video}
Kaituo Feng, Kaixiong Gong, Bohao Li, Zonghao Guo, Yibing Wang, Tianshuo Peng, Junfei Wu, Xiaoying Zhang, Benyou Wang, and Xiangyu Yue.
\newblock Video-r1: Reinforcing video reasoning in mllms.
\newblock \emph{arXiv preprint arXiv:2503.21776}, 2025.

\bibitem[Foss et~al.(2025)Foss, Evans, Mitts, Sinha, Rizvi, and Kao]{foss2025causalvqa}
Aaron Foss, Chloe Evans, Sasha Mitts, Koustuv Sinha, Ammar Rizvi, and Justine~T Kao.
\newblock Causalvqa: A physically grounded causal reasoning benchmark for video models.
\newblock \emph{arXiv preprint arXiv:2506.09943}, 2025.

\bibitem[Fu et~al.(2025)Fu, Dai, Luo, Li, Ren, Zhang, Wang, Zhou, Shen, Zhang, et~al.]{fu2025videomme}
Chaoyou Fu, Yuhan Dai, Yongdong Luo, Lei Li, Shuhuai Ren, Renrui Zhang, Zihan Wang, Chenyu Zhou, Yunhang Shen, Mengdan Zhang, et~al.
\newblock Video-mme: The first-ever comprehensive evaluation benchmark of multi-modal llms in video analysis.
\newblock In \emph{Proceedings of the Computer Vision and Pattern Recognition Conference}, pages 24108--24118, 2025.

\bibitem[Guo et~al.(2025)Guo, Yang, Zhang, Song, Zhang, Xu, Zhu, Ma, Wang, Bi, et~al.]{guo2025deepseek}
Daya Guo, Dejian Yang, Haowei Zhang, Junxiao Song, Ruoyu Zhang, Runxin Xu, Qihao Zhu, Shirong Ma, Peiyi Wang, Xiao Bi, et~al.
\newblock Deepseek-r1: Incentivizing reasoning capability in llms via reinforcement learning.
\newblock \emph{arXiv preprint arXiv:2501.12948}, 2025.

\bibitem[He et~al.(2025)He, Huang, Chen, Pei, Xu, Lu, and Pang]{he2025egoexobench}
Yuping He, Yifei Huang, Guo Chen, Baoqi Pei, Jilan Xu, Tong Lu, and Jiangmiao Pang.
\newblock Egoexobench: A benchmark for first-and third-person view video understanding in mllms.
\newblock \emph{arXiv preprint arXiv:2507.18342}, 2025.

\bibitem[Hern{\'a}n and Robins(2010)]{hernan2010causal}
Miguel~A Hern{\'a}n and James~M Robins.
\newblock Causal inference, 2010.

\bibitem[Hong et~al.(2025)Hong, Yu, Gu, Wang, Gan, Tang, Cheng, Qi, Ji, Pan, et~al.]{hong2025glm}
Wenyi Hong, Wenmeng Yu, Xiaotao Gu, Guo Wang, Guobing Gan, Haomiao Tang, Jiale Cheng, Ji Qi, Junhui Ji, Lihang Pan, et~al.
\newblock Glm-4.1 v-thinking: Towards versatile multimodal reasoning with scalable reinforcement learning.
\newblock \emph{arXiv e-prints}, pages arXiv--2507, 2025.

\bibitem[Lab(2025)]{qwen3vl2025}
Tongyi Lab.
\newblock Qwen3-vl: Large multimodal language models by alibaba cloud.
\newblock \url{https://huggingface.co/collections/Qwen/qwen3-vl}, 2025.
\newblock Model available at Hugging Face. Accessed: 2025-11-12.

\bibitem[Lai et~al.(2025)Lai, Li, Li, Liu, Li, and Zhao]{lai2025mini}
Xin Lai, Junyi Li, Wei Li, Tao Liu, Tianjian Li, and Hengshuang Zhao.
\newblock Mini-o3: Scaling up reasoning patterns and interaction turns for visual search.
\newblock \emph{arXiv preprint arXiv:2509.07969}, 2025.

\bibitem[Lei et~al.(2020)Lei, Yu, Berg, and Bansal]{lei2020tvqa+}
Jie Lei, Licheng Yu, Tamara Berg, and Mohit Bansal.
\newblock Tvqa+: Spatio-temporal grounding for video question answering.
\newblock In \emph{Proceedings of the 58th annual meeting of the association for computational linguistics}, pages 8211--8225, 2020.

\bibitem[Li et~al.(2022)Li, Niu, and Zhang]{li2022causalvidqa}
Jiangtong Li, Li Niu, and Liqing Zhang.
\newblock From representation to reasoning: Towards both evidence and commonsense reasoning for video question-answering.
\newblock In \emph{Proceedings of the IEEE/CVF conference on computer vision and pattern recognition}, pages 21273--21282, 2022.

\bibitem[Li et~al.(2023)Li, He, Wang, Li, Wang, Luo, Wang, Wang, and Qiao]{li2023videochat}
KunChang Li, Yinan He, Yi Wang, Yizhuo Li, Wenhai Wang, Ping Luo, Yali Wang, Limin Wang, and Yu Qiao.
\newblock Videochat: Chat-centric video understanding.
\newblock \emph{arXiv preprint arXiv:2305.06355}, 2023.

\bibitem[Li et~al.(2024{\natexlab{a}})Li, Wang, He, Li, Wang, Liu, Wang, Xu, Chen, Luo, et~al.]{li2024mvbench}
Kunchang Li, Yali Wang, Yinan He, Yizhuo Li, Yi Wang, Yi Liu, Zun Wang, Jilan Xu, Guo Chen, Ping Luo, et~al.
\newblock Mvbench: A comprehensive multi-modal video understanding benchmark.
\newblock In \emph{Proceedings of the IEEE/CVF Conference on Computer Vision and Pattern Recognition}, pages 22195--22206, 2024{\natexlab{a}}.

\bibitem[Li et~al.(2025)Li, Yan, Meng, Dong, Zeng, He, Wang, Qiao, Wang, and Wang]{li2025videochat}
Xinhao Li, Ziang Yan, Desen Meng, Lu Dong, Xiangyu Zeng, Yinan He, Yali Wang, Yu Qiao, Yi Wang, and Limin Wang.
\newblock Videochat-r1: Enhancing spatio-temporal perception via reinforcement fine-tuning.
\newblock \emph{arXiv preprint arXiv:2504.06958}, 2025.

\bibitem[Li et~al.(2024{\natexlab{b}})Li, Wang, and Jia]{li2024llama}
Yanwei Li, Chengyao Wang, and Jiaya Jia.
\newblock Llama-vid: An image is worth 2 tokens in large language models.
\newblock In \emph{European Conference on Computer Vision}, pages 323--340. Springer, 2024{\natexlab{b}}.

\bibitem[Maaz et~al.(2024)Maaz, Rasheed, Khan, and Khan]{maaz2024video}
Muhammad Maaz, Hanoona Rasheed, Salman Khan, and Fahad Khan.
\newblock Video-chatgpt: Towards detailed video understanding via large vision and language models.
\newblock In \emph{Proceedings of the 62nd Annual Meeting of the Association for Computational Linguistics (Volume 1: Long Papers)}, pages 12585--12602, 2024.

\bibitem[Meng et~al.(2025)Meng, Li, Wang, Tan, Zhang, Kong, Tong, Wang, Teng, Wang, et~al.]{meng2025openo3video}
Jiahao Meng, Xiangtai Li, Haochen Wang, Yue Tan, Tao Zhang, Lingdong Kong, Yunhai Tong, Anran Wang, Zhiyang Teng, Yujing Wang, et~al.
\newblock Open-o3 video: Grounded video reasoning with explicit spatio-temporal evidence.
\newblock \emph{arXiv preprint arXiv:2510.20579}, 2025.

\bibitem[OpenAI(2025)]{openai_gpt5_2025}
OpenAI.
\newblock Gpt-5.
\newblock \url{https://openai.com}, 2025.
\newblock Large language model.

\bibitem[Pearl(2009)]{pearl2009causal}
Judea Pearl.
\newblock Causal inference in statistics: An overview.
\newblock 2009.

\bibitem[Ravi et~al.(2024)Ravi, Gabeur, Hu, Hu, Ryali, Ma, Khedr, R{\"a}dle, Rolland, Gustafson, Mintun, Pan, Alwala, Carion, Wu, Girshick, Doll{\'a}r, and Feichtenhofer]{ravi2024sam2}
Nikhila Ravi, Valentin Gabeur, Yuan-Ting Hu, Ronghang Hu, Chaitanya Ryali, Tengyu Ma, Haitham Khedr, Roman R{\"a}dle, Chloe Rolland, Laura Gustafson, Eric Mintun, Junting Pan, Kalyan~Vasudev Alwala, Nicolas Carion, Chao-Yuan Wu, Ross Girshick, Piotr Doll{\'a}r, and Christoph Feichtenhofer.
\newblock Sam 2: Segment anything in images and videos.
\newblock \emph{arXiv preprint arXiv:2408.00714}, 2024.

\bibitem[Shi et~al.(2024)Shi, Di, Chen, and Xie]{shi2024agentdistill}
Yudi Shi, Shangzhe Di, Qirui Chen, and Weidi Xie.
\newblock Enhancing video-llm reasoning via agent-of-thoughts distillation.
\newblock \emph{arXiv preprint arXiv:2412.01694}, 2024.

\bibitem[Song et~al.(2024)Song, Chai, Wang, Zhang, Zhou, Wu, Chi, Guo, Ye, Zhang, et~al.]{song2024moviechat}
Enxin Song, Wenhao Chai, Guanhong Wang, Yucheng Zhang, Haoyang Zhou, Feiyang Wu, Haozhe Chi, Xun Guo, Tian Ye, Yanting Zhang, et~al.
\newblock Moviechat: From dense token to sparse memory for long video understanding.
\newblock In \emph{Proceedings of the IEEE/CVF Conference on Computer Vision and Pattern Recognition}, pages 18221--18232, 2024.

\bibitem[Su et~al.(2025)Su, Wang, Ren, Lin, and Chen]{su2025pixel}
Alex Su, Haozhe Wang, Weiming Ren, Fangzhen Lin, and Wenhu Chen.
\newblock Pixel reasoner: Incentivizing pixel-space reasoning with curiosity-driven reinforcement learning.
\newblock \emph{arXiv preprint arXiv:2505.15966}, 2025.

\bibitem[Wang et~al.(2025{\natexlab{a}})Wang, Li, Huang, Wang, Wang, Zhang, Zheng, Bai, Kang, Feng, et~al.]{wang2025traceable}
Haochen Wang, Xiangtai Li, Zilong Huang, Anran Wang, Jiacong Wang, Tao Zhang, Jiani Zheng, Sule Bai, Zijian Kang, Jiashi Feng, et~al.
\newblock Traceable evidence enhanced visual grounded reasoning: Evaluation and methodology.
\newblock \emph{arXiv preprint arXiv:2507.07999}, 2025{\natexlab{a}}.

\bibitem[Wang et~al.(2025{\natexlab{b}})Wang, Gao, Gu, Pu, Cui, Wei, Liu, Jing, Ye, Shao, et~al.]{wang2025internvl3_5}
Weiyun Wang, Zhangwei Gao, Lixin Gu, Hengjun Pu, Long Cui, Xingguang Wei, Zhaoyang Liu, Linglin Jing, Shenglong Ye, Jie Shao, et~al.
\newblock Internvl3.5: Advancing open-source multimodal models in versatility, reasoning, and efficiency.
\newblock \emph{arXiv preprint arXiv:2508.18265}, 2025{\natexlab{b}}.

\bibitem[Wang et~al.(2024)Wang, Zeng, Zheng, Xing, Xu, and Xu]{wang2024videocot}
Yan Wang, Yawen Zeng, Jingsheng Zheng, Xiaofen Xing, Jin Xu, and Xiangmin Xu.
\newblock Videocot: A video chain-of-thought dataset with active annotation tool.
\newblock \emph{arXiv preprint arXiv:2407.05355}, 2024.

\bibitem[Wang et~al.(2025{\natexlab{c}})Wang, Wang, Xu, Du, Lin, Xiao, Yue, Ju, Zhang, Yang, et~al.]{wang2025time}
Ye Wang, Ziheng Wang, Boshen Xu, Yang Du, Kejun Lin, Zihan Xiao, Zihao Yue, Jianzhong Ju, Liang Zhang, Dingyi Yang, et~al.
\newblock Time-r1: Post-training large vision language model for temporal video grounding.
\newblock \emph{arXiv preprint arXiv:2503.13377}, 2025{\natexlab{c}}.

\bibitem[Xiao et~al.(2021)Xiao, Shang, Yao, and Chua]{xiao2021next}
Junbin Xiao, Xindi Shang, Angela Yao, and Tat-Seng Chua.
\newblock Next-qa: Next phase of question-answering to explaining temporal actions.
\newblock In \emph{Proceedings of the IEEE/CVF conference on computer vision and pattern recognition}, pages 9777--9786, 2021.

\bibitem[Xiao et~al.(2024)Xiao, Yao, Li, and Chua]{xiao2024nextgqa}
Junbin Xiao, Angela Yao, Yicong Li, and Tat-Seng Chua.
\newblock Can i trust your answer? visually grounded video question answering.
\newblock In \emph{Proceedings of the IEEE/CVF Conference on Computer Vision and Pattern Recognition}, pages 13204--13214, 2024.

\bibitem[Xiaomi(2025)]{coreteam2025mimovltechnicalreport}
LLM-Core-Team Xiaomi.
\newblock Mimo-vl technical report, 2025.

\bibitem[Xu et~al.(2017)Xu, Zhao, Xiao, Wu, Zhang, He, and Zhuang]{xu2017video}
Dejing Xu, Zhou Zhao, Jun Xiao, Fei Wu, Hanwang Zhang, Xiangnan He, and Yueting Zhuang.
\newblock Video question answering via gradually refined attention over appearance and motion.
\newblock In \emph{Proceedings of the 25th ACM international conference on Multimedia}, pages 1645--1653, 2017.

\bibitem[Xu et~al.(2025)Xu, Li, Zhou, Wan, Zhang, Korhonen, and Vuli{\'c}]{xu2025visual}
Yi Xu, Chengzu Li, Han Zhou, Xingchen Wan, Caiqi Zhang, Anna Korhonen, and Ivan Vuli{\'c}.
\newblock Visual planning: Let's think only with images.
\newblock \emph{arXiv preprint arXiv:2505.11409}, 2025.

\bibitem[Yu et~al.(2025)Yu, Wu, Chu, Ren, Huang, Chu, Zhang, He, Li, Li, et~al.]{yu2025vrbench}
Jiashuo Yu, Yue Wu, Meng Chu, Zhifei Ren, Zizheng Huang, Pei Chu, Ruijie Zhang, Yinan He, Qirui Li, Songze Li, et~al.
\newblock Vrbench: A benchmark for multi-step reasoning in long narrative videos.
\newblock \emph{arXiv preprint arXiv:2506.10857}, 2025.

\bibitem[Zang et~al.(2023)Zang, Wang, Pei, and Liang]{zang2023discovering}
Chuanqi Zang, Hanqing Wang, Mingtao Pei, and Wei Liang.
\newblock Discovering the real association: Multimodal causal reasoning in video question answering.
\newblock In \emph{Proceedings of the IEEE/CVF Conference on Computer Vision and Pattern Recognition}, pages 19027--19036, 2023.

\bibitem[Zhang et~al.(2025{\natexlab{a}})Zhang, Gu, Li, Ma, Bai, Zhang, Zhang, Zhou, He, and Tang]{zhang2025thinking}
Haoji Zhang, Xin Gu, Jiawen Li, Chixiang Ma, Sule Bai, Chubin Zhang, Bowen Zhang, Zhichao Zhou, Dongliang He, and Yansong Tang.
\newblock Thinking with videos: Multimodal tool-augmented reinforcement learning for long video reasoning.
\newblock \emph{arXiv preprint arXiv:2508.04416}, 2025{\natexlab{a}}.

\bibitem[Zhang et~al.(2025{\natexlab{b}})Zhang, Hao, Tang, Zhang, Wang, Wang, Ma, and Zhang]{zhang2025video}
Shuyi Zhang, Xiaoshuai Hao, Yingbo Tang, Lingfeng Zhang, Pengwei Wang, Zhongyuan Wang, Hongxuan Ma, and Shanghang Zhang.
\newblock Video-cot: A comprehensive dataset for spatiotemporal understanding of videos based on chain-of-thought.
\newblock In \emph{Proceedings of the 33rd ACM International Conference on Multimedia}, pages 12745--12752, 2025{\natexlab{b}}.

\bibitem[Zhang et~al.(2025{\natexlab{c}})Zhang, Gao, Zhang, Li, Zhang, Liu, Yuan, Wu, Jia, Zhu, et~al.]{zhang2025chain}
Xintong Zhang, Zhi Gao, Bofei Zhang, Pengxiang Li, Xiaowen Zhang, Yang Liu, Tao Yuan, Yuwei Wu, Yunde Jia, Song-Chun Zhu, et~al.
\newblock Chain-of-focus: Adaptive visual search and zooming for multimodal reasoning via rl.
\newblock \emph{arXiv preprint arXiv:2505.15436}, 2025{\natexlab{c}}.

\bibitem[Zhang et~al.(2025{\natexlab{d}})Zhang, Wen, Wu, and Huang]{zhang2025tinyllava}
Xingjian Zhang, Siwei Wen, Wenjun Wu, and Lei Huang.
\newblock Tinyllava-video-r1: Towards smaller lmms for video reasoning.
\newblock \emph{arXiv preprint arXiv:2504.09641}, 2025{\natexlab{d}}.

\bibitem[Zhang et~al.(2024)Zhang, Li, Liu, Lee, Gui, Fu, Feng, Liu, and Li]{zhang2024llavanextvideo}
Yuanhan Zhang, Bo Li, haotian Liu, Yong~jae Lee, Liangke Gui, Di Fu, Jiashi Feng, Ziwei Liu, and Chunyuan Li.
\newblock Llava-next: A strong zero-shot video understanding model, 2024.

\bibitem[Zhang et~al.(2020)Zhang, Zhao, Zhao, Wang, Liu, and Gao]{zhang2020videostg}
Zhu Zhang, Zhou Zhao, Yang Zhao, Qi Wang, Huasheng Liu, and Lianli Gao.
\newblock Where does it exist: Spatio-temporal video grounding for multi-form sentences.
\newblock In \emph{Proceedings of the IEEE/CVF Conference on Computer Vision and Pattern Recognition}, pages 10668--10677, 2020.

\bibitem[Zheng et~al.(2024)Zheng, Zhang, Zhang, Ye, Luo, Feng, and Ma]{zheng2024llamafactory}
Yaowei Zheng, Richong Zhang, Junhao Zhang, Yanhan Ye, Zheyan Luo, Zhangchi Feng, and Yongqiang Ma.
\newblock Llamafactory: Unified efficient fine-tuning of 100+ language models.
\newblock In \emph{Proceedings of the 62nd Annual Meeting of the Association for Computational Linguistics (Volume 3: System Demonstrations)}, Bangkok, Thailand, 2024. Association for Computational Linguistics.

\bibitem[Zheng et~al.(2025)Zheng, Yang, Hong, Zhao, Xu, Yang, Shen, and Yu]{zheng2025deepeyes}
Ziwei Zheng, Michael Yang, Jack Hong, Chenxiao Zhao, Guohai Xu, Le Yang, Chao Shen, and Xing Yu.
\newblock Deepeyes: Incentivizing" thinking with images" via reinforcement learning.
\newblock \emph{arXiv preprint arXiv:2505.14362}, 2025.

\end{thebibliography}
}

\clearpage
\setcounter{page}{1}
\maketitlesupplementary

\section{Appendix Overview}
\label{sec:supp_overview}

The organization of the appendix is as follows:

\cref{sec:supp_qa_example}: CaST-Bench Samples from Each Category

\cref{sec:supp_pipeline}: Details of Data Annotation Pipeline

\cref{sec:supp_experiment_setup}: Details of Experiment Setup

\cref{sec:supp_failure}: Case Studies and Failure Analysis

\cref{sec:supp_social}: Social Impact, License, and Access

\section{CaST-Bench Samples from Each Category}
\label{sec:supp_qa_example}

Due to page limitation in the main manuscript, we show more examples regarding all question types, as follows.

\begin{figure*}
    \centering
    \includegraphics[width=0.99\linewidth]{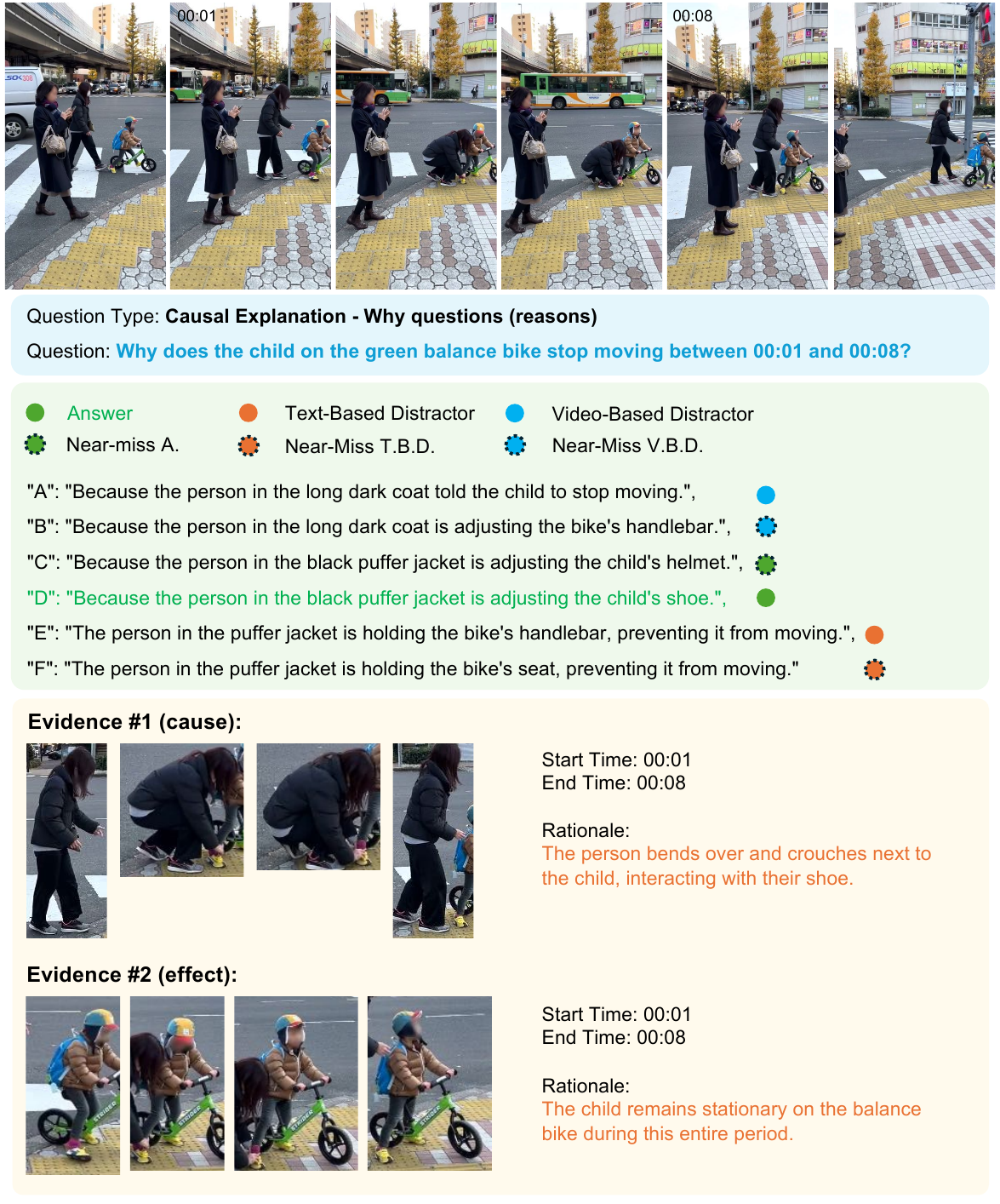}
    \caption{\textbf{CaST-Bench Data Sample 1.} Question Type: Causal Explanation - Why questions (reasons).}
    \label{fig:supp_sample_1}
\end{figure*}

\begin{figure*}
    \centering
    \includegraphics[width=0.99\linewidth]{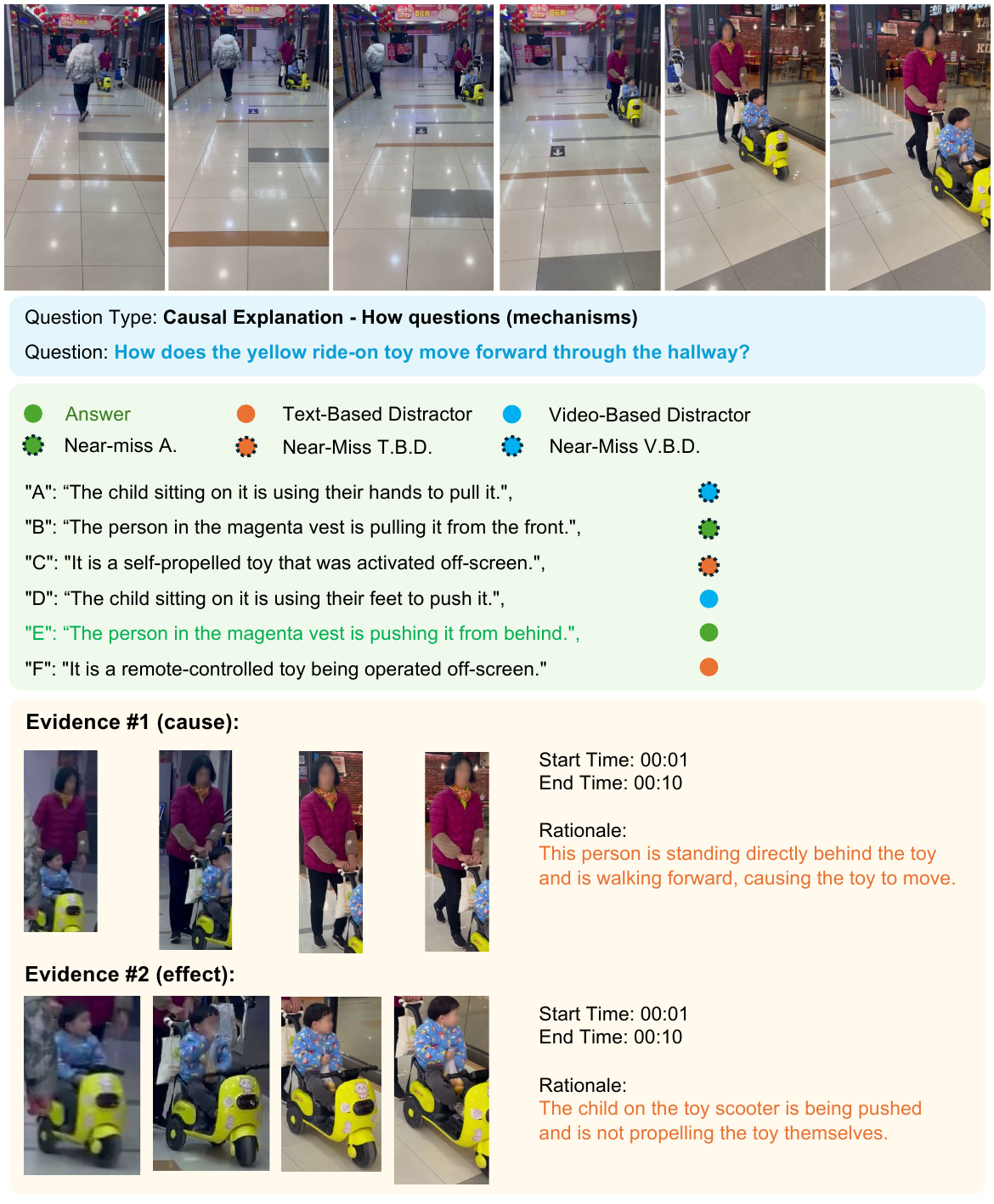}
    \caption{\textbf{CaST-Bench Data Sample 2.} Question Type: Causal Explanation - How questions (mechanisms).}
    \label{fig:supp_sample_2}
\end{figure*}

\begin{figure*}
    \centering
    \includegraphics[width=0.99\linewidth]{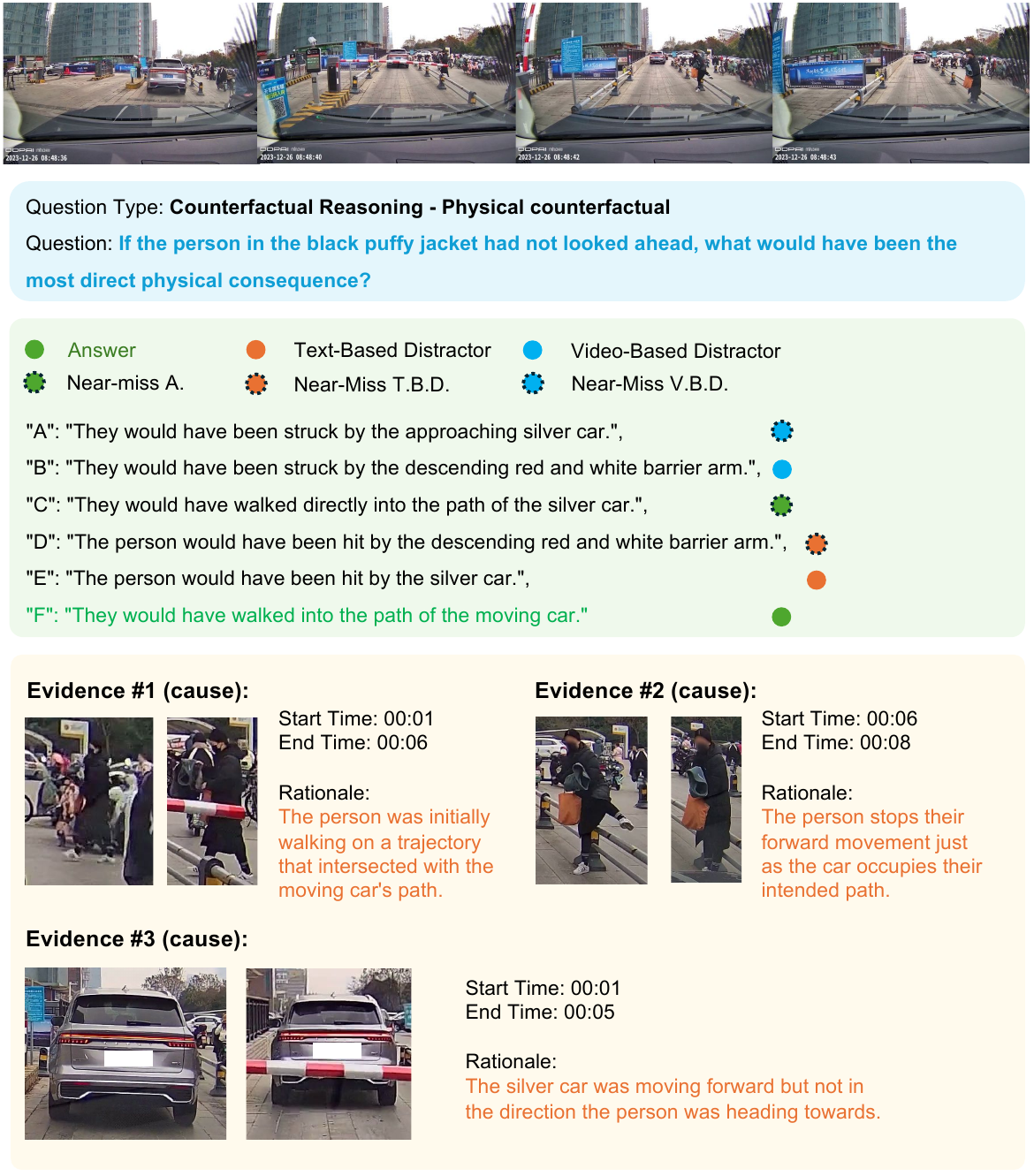}
    \caption{\textbf{CaST-Bench Data Sample 3.} Question Type: Counterfactual Reasoning - Physical counterfactual.}
    \label{fig:supp_sample_3}
\end{figure*}

\begin{figure*}
    \centering
    \includegraphics[width=0.99\linewidth]{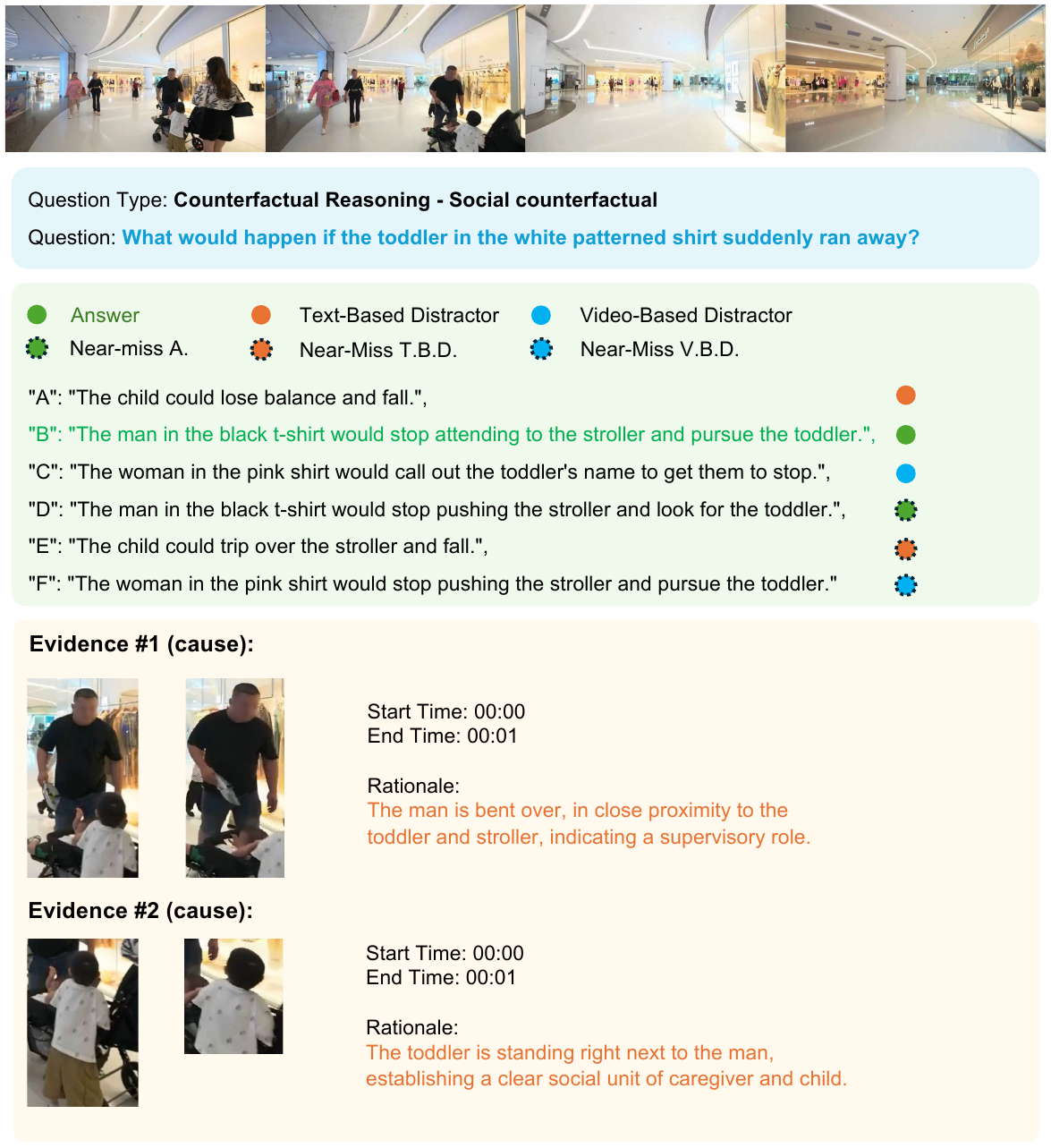}
    \caption{\textbf{CaST-Bench Data Sample 4.} Question Type: Counterfactual Reasoning - Physical counterfactual.}
    \label{fig:supp_sample_4}
\end{figure*}

\begin{figure*}
    \centering
    \includegraphics[width=0.99\linewidth]{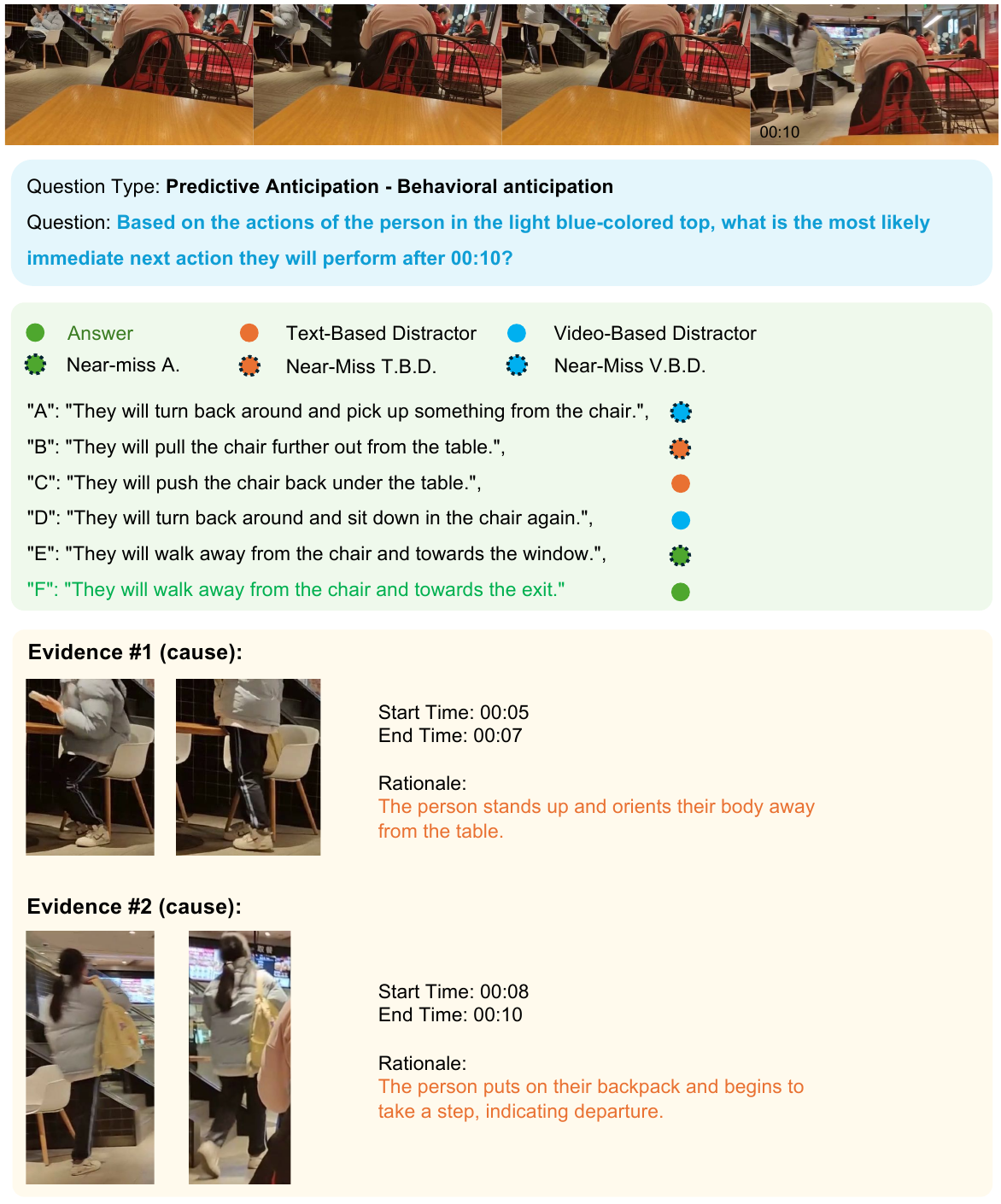}
    \caption{\textbf{CaST-Bench Data Sample 5.} Question Type: Counterfactual Reasoning - Social counterfactual.}
    \label{fig:supp_sample_5}
\end{figure*}

\begin{figure*}
    \centering
    \includegraphics[width=0.99\linewidth]{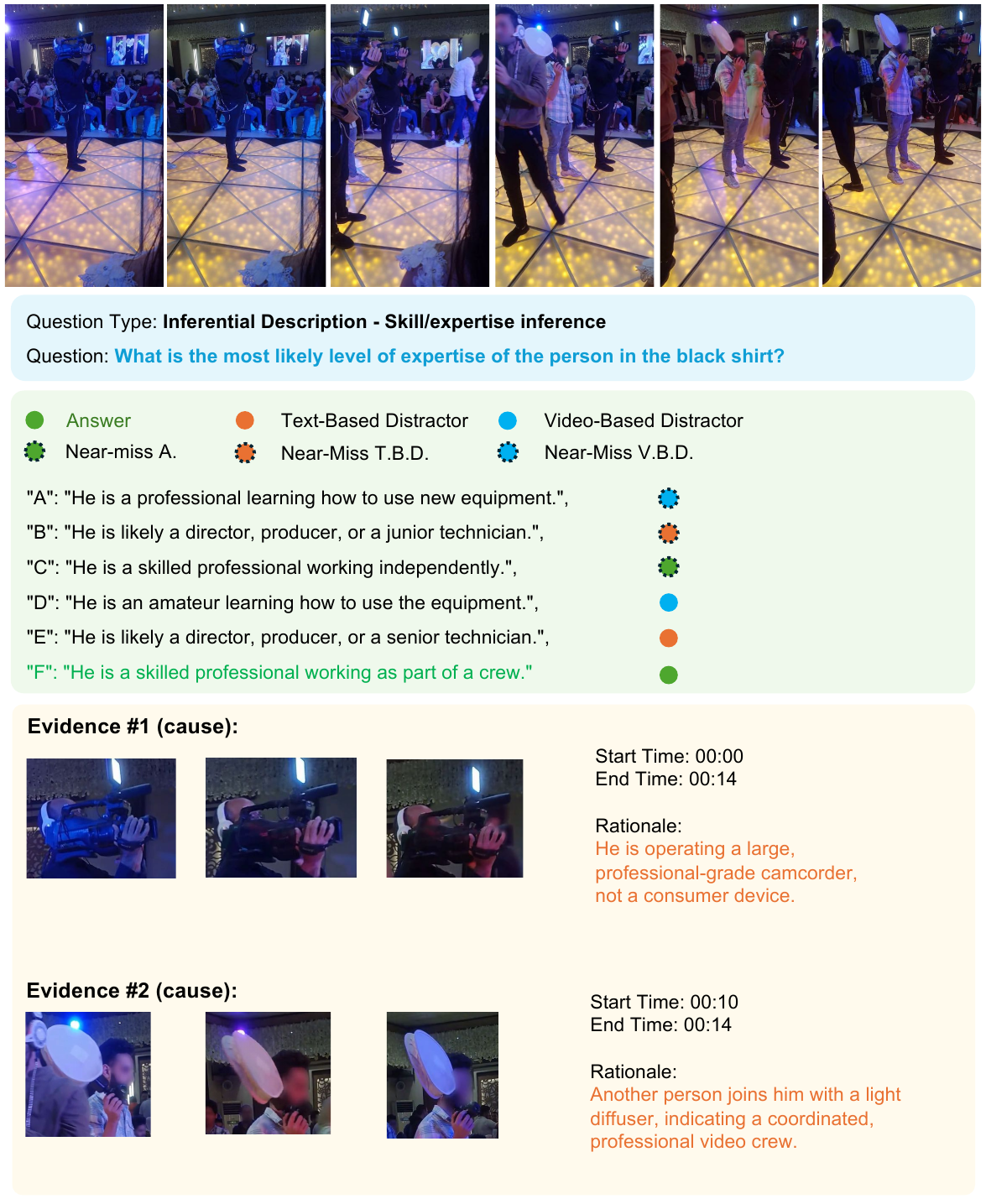}
    \caption{\textbf{CaST-Bench Data Sample 6.} Question Type: Predictive Anticipation - Behavioral anticipation.}
    \label{fig:supp_sample_6}
\end{figure*}

\begin{itemize}
    \item \textbf{Causal Explanation} Questions that explain the reasons (\textit{why}) or mechanisms (\textit{how}) behind actions or events.
    \begin{itemize}
        \item Why questions (reasons), shown in \cref{fig:supp_sample_1}
        \item How questions (mechanisms), shown in \cref{fig:supp_sample_2}
    \end{itemize}
    \item \textbf{Counterfactual Reasoning} Questions that infer the consequence of a single alteration to a key causal element.
    \begin{itemize}
        \item Physical counterfactual, shown in \cref{fig:supp_sample_3}
        \item Social counterfactual, shown in \cref{fig:supp_sample_4}
    \end{itemize}
    \item \textbf{Predictive Anticipation} Questions that require predicting the most probable and immediate outcome.
    \begin{itemize}
        \item Behavioral anticipation, shown in \cref{fig:supp_sample_5}
    \end{itemize}
    \item \textbf{Inferential Description} Questions that infer implicit attributes or states (e.g., roles, intentions, emotions).
    \begin{itemize}
        \item Skill/expertise inference, shown in \cref{fig:supp_sample_6}
        \item ... 
    \end{itemize}
\end{itemize}

Detailed definition of each question type can be found in \cref{sec:supp_pipeline_qa_and_causal}.

\section{Details of Data Annotation Pipeline}
\label{sec:supp_pipeline}

\subsection{Video Selection}
\label{sec:supp_pipeline_video_selection}

Our benchmark targets causal reasoning in realistic, cluttered scenes where multiple actors interact over time. As discussed in \cref{sec:cast_bench}, carefully curating the raw video pool is essential: studio footage or single-actor clips often lack the competing causal cues and spatial ambiguity needed to stress-test grounding. We therefore begin CaST-Bench construction with a \textit{Video Selection} stage that selects complex activities from the SegmentAnything-Video (SAV) dataset~\cite{ravi2024sam2}, ensuring that downstream annotation stages receive videos containing diverse objects, overlapping motions, and stable visual quality.

We operationalize the above motivation with a tool that iterates over SAV clips matching a curator-specified pattern, submits each video to a frontier multimodal model via a standard API endpoint, and relies on the prompt (shown below) to approve a clip only when (a) multiple humans are visible and (b) the visible people are engaged in different actions, which directly encourages multi-instance causal chains; the same prompt rejects clips with motion artifacts (camera shake, rapid cuts or pans), low visibility, or severe compression because poor spatial alignment would confound later evidence annotation, and the resulting 1{,}015 accepted videos constitute the seed pool for the remaining stages of the pipeline.

\begin{tcolorbox}[
    colback=gray!15,
    colframe=gray!50,
    enhanced,
    arc=6pt,
    boxrule=1pt,
    title={Prompt 1: Automated Video Selection},
    fonttitle=\bfseries,
    colbacktitle=gray!40,
    coltitle=black,
    left=5pt,right=5pt,top=5pt,bottom=5pt
]
\footnotesize
You are a video analysis expert.
Task
- Watch the entire video.
Return exactly one of:
- True — the video shows multiple people engaged in different actions.
- False — otherwise.
Output format
- Return only True or False (no quotes, no extra text, no punctuation).
Decision rules
1) Human presence
- If only one person is present, return False.
- If multiple people are present and each visible person's behavior are differnt, return True.
2) Visual quality and stability
- Return False if there is:
  a) strong camera shake;
  b) rapid viewpoint shifts or frequent cuts;
  c) very low image quality (subject not discernible, heavy blur, severe compression, lighting too dark or overexposed);
  d) fast camera movement (e.g., quick or whip pans);
  e) severe camera tilt.
Remember: return only True or False.
\end{tcolorbox}

\subsection{Spatio-Temporal Fine-grained Instance Description}
\label{sec:supp_pipeline_spatio_temporal}

Generating detailed and accurate descriptions for specific object instances within complex, dynamic scenes presents a significant challenge. Vision-Language Models (VLMs) can be easily distracted by background clutter or other moving objects, leading to descriptions that are either inaccurate or lack focus on the intended instance. To address this, we developed a multi-stage pipeline that systematically isolates each instance and generates both static and dynamic descriptions, forming a rich corpus for the subsequent generation of causal questions. This process consists of three key stages: instance isolation for dynamic analysis, static description for contextual understanding, and dynamic description for capturing temporal behavior.

\paragraph{Stage 1: Instance Isolation for Dynamic Analysis}
The primary challenge in analyzing the behavior of a single object is isolating it from the surrounding environment. To enable a VLM to focus exclusively on the actions of one instance, we first create a cropped video clip that follows the target object throughout its appearance in the original video. Using the instance masks from the SAV dataset, our pipeline calculates the bounding box for the target instance in each frame and generates a new, smaller video. This clip is padded to a uniform size and dynamically re-centered in each frame to ensure the instance remains the focus of the shot. This isolation is crucial for the final stage of dynamic description, as it provides the VLM with an unambiguous view of the instance's actions, free from the interference of other objects or background events.

\begin{figure*}[t]
    \label{prompt2:inst_description}
    \footnotesize
    \begin{tcolorbox}[
        colback=gray!15,
        colframe=gray!50,
        enhanced,
        arc=6pt,
        boxrule=1pt,
        title={Prompt 2: Static Instance Description},
        fonttitle=\bfseries,
        colbacktitle=gray!40,
        coltitle=black,
        left=5pt,right=5pt,top=5pt,bottom=5pt
    ]
    You are given two aligned images of the same scene:
    
    1) The original image where the target instance is marked with a green outline (the outline is an overlay, not part of the object). 2) A version of the original image where the background is blurred to isolate the target instance.
    
    Objective:
    
    Write exactly one English description that refers only to the target instance and its scene/context in the original image.
    
    Strict Rules:
    
    * Focus: Describe only the focused object (the sharp item with green outline); ignore all other discrete objects.
    
    * Content: If visible, include the object's visual attributes, function, or role, and its environmental context.
    
    * Grammar: Do not use verbs or describe dynamic actions; use only nouns, adjectives, and prepositional phrases; structure around a single primary object noun phrase.
    
    * People: Do not describe any person's actions or body poses.
    
    * Exclusions: Do not mention the green outline, background blur, black paddings, image cropping, or other image artifacts; do not include quantities or numbers; do not describe the positional relationship between the camera and the target instance (e.g. from behind).
    
    * Color Determination: Identify color only from the object's internal surfaces and textures, excluding any borders or edges affected by overlays. If the true color is uncertain, omit color adjectives.
    
    * Outline Ignorance: Never describe the object as "green" or assign any color due to the green outline; the outline must be ignored for all attributes. 
    
    * Style: Objective, specific, and under 30 words.
    
    * Validity: If any object other than the focused instance is described, the output is invalid.
    
    Output Format:
    
    * Exactly one sentence, under 30 words.
    
    * Begin with a single generic category noun in brackets, followed by a colon and the description.
    
    * Format: "[noun]: description."
    \end{tcolorbox}
    
    \begin{tcolorbox}[
        colback=gray!15,
        colframe=gray!50,
        enhanced,
        arc=6pt,
        boxrule=1pt,
        title={Prompt 3: Dynamic Instance Description},
        fonttitle=\bfseries,
        colbacktitle=gray!40,
        coltitle=black,
        left=5pt,right=5pt,top=5pt,bottom=5pt
    ]
    You are an AI video analysis model specializing in tracking and describing the dynamics of a single object over time. 
    
    You receive two inputs:
    
    1. **Text Description**: A sentence identifying the target object and its surrounding scene and context.
    
    2. **Video Clip**: A silent video focused on the target object. The object is highlighted with a green border for tracking purposes.
    
    **Task**: Generate a time-stamped log detailing the specific dynamics of the specified object shown in the video.
    
    **Rules**:
    
    - Source of Truth: The video clip is the source of truth. The text input is for context only. If a frame is completely black, it signifies the object is absent from the original footage during that time and do not create an entry describing the absence.
    
    - Focus on Dynamics: Describe the object's actions, movements, transformations in state, interactions with others, and group activities in detail. Do not describe static properties if the object remains unchanged. If the object is motionless for the entire video, provide a single entry for the full duration.
    
    - No Direction: Do not describe movement directions (e.g., left/right/up/down).
    
    - Ignore Occlusion and Camera Motion Artifacts: Describe only changes inherent to the object itself or caused by direct interaction. Do not describe visual changes (such as fragmentation) caused by foreground occlusion and camera motion (e.g., the object being partially hidden by something passing in front of it or the object moves out of frame becuase of the camera motion).
    
    - Camera vs Object Motion: Distinguish camera motion from object motion. Do not attribute camera movement to the object. Describe only the object's movement; do not describe camera movement.
    
    - Ignore Visual Aids: Do not mention the green border lines and black paddings in the description.
    
    - Consolidate Time: Merge continuous periods of the same action or inaction into a single time-stamped entry.
    
    **Strict Output Format** (no extra commentary or sentences):
    
    '[MM:SS] - [MM:SS]: [instance dynamics description]'
    \end{tcolorbox}
\end{figure*}

\paragraph{Stage 2: Static Description for Contextual Understanding}
To capture essential contextual information, we generate a static description based on the full, uncropped video frame. For each instance, we identify a single ``best'' representative frame from the original video, typically one where the object is clearly visible and centrally located. 

We then present two images to the VLM: (1) the original, full-resolution frame with the target instance highlighted by a green outline, and (2) the same frame but with the background heavily blurred, further emphasizing the instance. This dual-image input allows the model to observe the object within its complete environment, facilitating a more holistic understanding. The VLM is then prompted to provide a concise, single-sentence description of the object's appearance and its role within the scene.

The prompt used for this stage is shown \hyperref[prompt2:inst_description]{Prompt 2}.

\paragraph{Stage 3: Dynamic Description for Capturing Temporal Behavior}
With a static, contextual description in hand, the final stage is to capture the instance's dynamic behavior over time. For this, we use the cropped video clip generated in Stage 1. This focused input prevents the model from being distracted by other scene elements. We provide the VLM with both the cropped video and the static description generated in Stage 2. The static description acts as a crucial anchor, informing the model about \textit{what} the object is, while the video shows \textit{what the object does}. The model is prompted to produce a detailed, time-stamped log of the object's movements, actions, and changes in state, effectively creating a fine-grained temporal narrative for each instance. These detailed, dynamic descriptions form the basis from which our causal questions and evidence chains are generated.

The prompt used for this stage is shown in \hyperref[prompt2:inst_description]{Prompt 3}.

\begin{figure*}[t]
    \centering
    \includegraphics[width=0.99\linewidth]{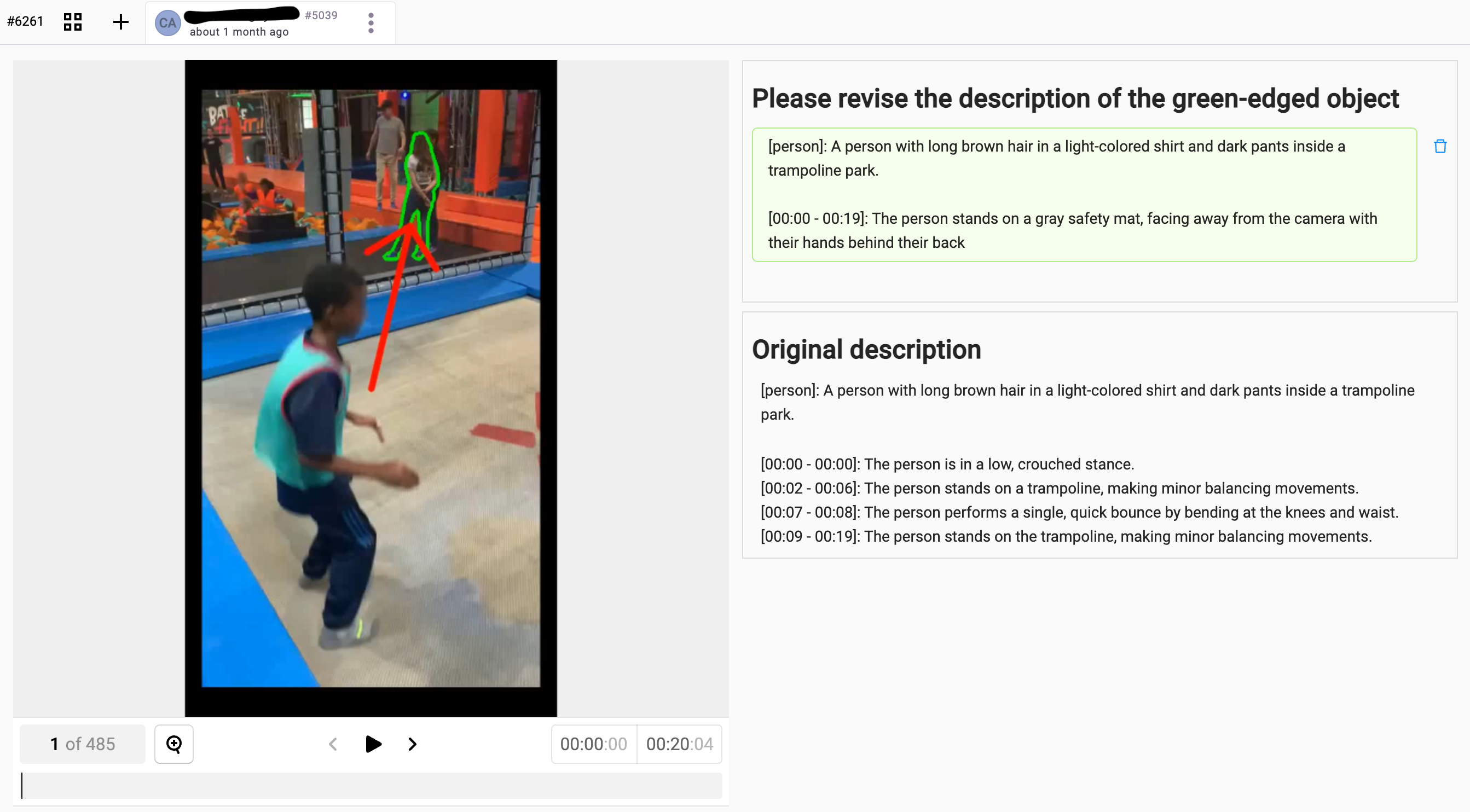}
    \caption{\textbf{Interface of Human Annotation for Instance Description (\cref{sec:supp_pipeline_human_annotation}).} The left part displays the video, and on the right there is a text box where the annotator can freely edit or modify the content, based on original description that is AI-generated (\cref{sec:supp_pipeline_spatio_temporal}).}
    \label{fig:supp_webpage}
\end{figure*}

\subsection{Human Annotation for Instance Description}
\label{sec:supp_pipeline_human_annotation}

The AI-generated captions from \cref{sec:supp_pipeline_spatio_temporal} provide dense coverage for every tracked instance, yet they remain vulnerable to the very issues CaST-Bench is designed to surface: cluttered scenes, tiny objects, and subtle causal cues often mislead frontier VLMs. To avoid propagating these hallucinations into later QA generation stages, we added a dedicated human verification pass that rewrites each instance description so that it is temporally faithful to the underlying video and linguistically precise.

\paragraph{Interface and Workflow}
We built a custom web interface (see \cref{fig:supp_webpage}) that streams the source video on the left panel while highlighting the target instance with the same green edge used throughout the pipeline. The right panel hosts an editable textbox seeded with the VLM-produced, time-stamped description plus controls for adding or removing timeline rows. Annotators proceed in two steps: (1) watch the full clip to understand the object's behavior, and (2) iteratively edit the draft description until every time span and action aligns with the video evidence. This tight coupling between visualization and editing allows annotators to immediately localize contradictions and keeps the overall ST formatting consistent with downstream tooling.

\paragraph{Annotation Guidelines}
We distilled the guideline into three concrete requirements. \textit{(i) Contradiction removal}: every sentence must be cross-checked against the clip; any mismatch between the AI text and the video, including incorrect attributes, missed interactions, or spurious events, must be corrected before submission. \textit{(ii) Temporal correctness}: if misaligned time breaks cause errors, annotators are instructed to split, merge, add, or delete timeline rows while preserving the canonical ``[start--end]: action'' format so that causal evidence extraction remains deterministic. \textit{(iii) Special cases}: when the interface shows an ``Absent'' frame, annotators leave the corresponding time range empty; if the instance is too small or blurred to describe reliably, they either click ``Skip'' or enter ``null'' to flag the sample. These rules ensure that only verifiable, grounded content flows into the causal chain generator.

\paragraph{Annotator Expertise and Quality Control}
We recruited seven professional annotators who are native English speakers to maximize both factual accuracy and grammatical fluency. Each annotator underwent a calibration phase using the written handbook accompanying the interface, during which we sampled revisions for spot checks and provided feedback on missed contradictions. The combination of expert annotators, a purpose-built tool, and strict revision policies yields 10{,}728 timestamp-accurate instance descriptions that reliably serve as seed evidence for CaST-Bench.

\subsection{QA and Causal Chain Generation with Explicit ST Evidences}
\label{sec:supp_pipeline_qa_and_causal}

\paragraph{Motivation and Challenges}
After the human-verified instance narratives in \cref{sec:supp_pipeline_human_annotation}, the next bottleneck is converting thousands of time-stamped descriptions into causal QA items whose answers are only recoverable when a model reconstructs the same spatio-temporal chain. The scenes we curate often contain multiple overlapping human-object interactions, so naively prompting a VLM frequently yields generic ``what'' questions that ignore the causal cues we need. Furthermore, even when high-quality QA proposals emerge, they must be expressed in a structured format that later stages (distractor generation, mask-based validation, metric computation) can parse without manual post-processing. Finally, each textual evidence snippet must be tied back to the exact pixel tracks that human annotators produced; otherwise the benchmark would fail to measure grounded reasoning. The pipeline described below addresses these pain points in tightly coupled steps.

\paragraph{Step 1. Consolidating instance evidence into a causal reasoning canvas}
For every SAV video directory, we scan all verified caption JSONs, skip any file explicitly flagged as unusable, and extract the instance identifier together with its static appearance sentence and dynamic timeline. A normalization routine merges the static and dynamic sentences into a single ``Combined Video Descriptions'' block (as shown in \hyperref[prompt4:Causal_QA_Generation_Prompt]{Prompt 4}) where each entry begins with the instance identifier followed by the appearance rationale and the chronological action log. This formatted block is the only textual context the downstream VLM sees; it forces the model to reason over causal candidates spanning different people, tools, and time ranges.

\paragraph{Step 2. Prompting the model as a causal QA author}
The generation step issues a single, large instruction to a video-aware language model. The instruction injects five ingredients: (i) the combined descriptions, (ii) the raw video, (iii) the causal question taxonomy (\hyperref[prompt6:Causal_Question_Taxonomy]{Prompt 6}), (iv) a JSON schema (\hyperref[prompt5:qa_evidence_template]{Prompt 5}) that the output must obey. The model is required to return \emph{exactly} six QA-Evidence items at once so that diversity is enforced within a single decoding call, and the string formatter verifies that every placeholder in the prompt has been replaced before the request is sent. The exact instructions are reproduced in \hyperref[prompt4:Causal_QA_Generation_Prompt]{Prompt 4}).

\paragraph{Step 3. Enforcing a machine-checkable schema}
Even with a careful prompt, large models sometimes drift away from the desired structure. We therefore include the authoritative JSON layout in every call and validate the response immediately after decoding. Any malformed response is wrapped into an ``error'' JSON so that the dataset builders can retry the sample later. The schema is shown in \hyperref[prompt5:qa_evidence_template]{Prompt 5}; note how every field explicitly asks for temporal boundaries and instance identifiers, which later enable automatic alignment with bounding boxes.

\paragraph{Step 4. Covering the full causal taxonomy}
We embed the entire set of question types inside the prompt and instruct the model to cycle through subcategories when possible. This design choice directly supports the taxonomy introduced in the main paper (\cref{sec:cast_bench}) and ensures that the resulting benchmark stresses explanation, counterfactual, prediction, and inference evenly. The taxonomy text shipped to the model is given at \hyperref[prompt6:Causal_Question_Taxonomy]{Prompt 6}.

\paragraph{Step 5. Injecting explicit spatio-temporal evidence}
Once a valid JSON exists, a lightweight post-processor takes the question subject and every evidence instance, parses their reported start/end times in \texttt{mm:ss} format, and converts them into integer-second indices. It then loads the dense bounding-box tracks collected earlier (one JSON per tracked instance) and slices the corresponding coordinates for each second within the requested range. Two augmentations are written back into the QA structure: (i) a stable reference to the source track file (``SAV\_instance\_id'') and (ii) a dictionary named ``bboxes\_in\_range'' that maps every aligned second to the \([x_{\min}, y_{\min}, x_{\max}, y_{\max}]\) box recorded at 1~FPS.

\begin{figure*}
\label{prompt4:Causal_QA_Generation_Prompt}
\begin{tcolorbox}[
    colback=gray!15,
    colframe=gray!50,
    enhanced,
    arc=6pt,
    boxrule=1pt,
    title={Prompt 4: Causal QA Generation},
    fonttitle=\bfseries,
    colbacktitle=gray!40,
    coltitle=black,
    left=5pt,right=5pt,top=5pt,bottom=5pt
]
\footnotesize
You are a video analysis expert specializing in causal reasoning. Watch the video and read the time-stamped, instance-level descriptions, and then generate six high-quality causal QA-Evidence items. The output must strictly follow the provided JSON template.
\newline

INPUTS:

- Original video
- Combined Video Descriptions (instance-level descriptions with timestamps)
- Causal Questions Taxonomy
- QA-Evidence Template (authoritative schema)
- Example QA-Evidences
\newline

Combined Video Descriptions:
\{captions\_text\}
\newline

Causal Questions Taxonomy
\{question\_types\}
\newline

QA-Evidence Template
\{qa\_evidence\_template\}
\newline

TASK:

Create six QA-Evidence following subcategories in the Causal Questions Taxonomy. Produce a single JSON object that exactly matches the QA-Evidence Template, with top-level key 'QA\_Evidence\_List' containing exactly six QA items.
\newline

FIELD SPECIFICATION (align with the QA-Evidence Template):

- Question
  - question\_text: The question text. Do not include any instance\_id.
  - subject\_instance\_id: if the question\_text is asking about a specific subject, find the corresponding instance\_id; otherwise null.
  - question\_start\_time \& question\_end\_time: if the question\_text mentions a specific time period, write "mm:ss"; otherwise null.
  - question\_type\_subcategory: one subcategory number from Causal Questions Taxonomy
  - question\_type\_subcategory\_name: exact name of the subcategory
  - video\_scene\_category: Concise scene label to describe the scene.
- Answer
  - answer\_text: within 15 words. Do not include any instance\_id.
- Evidence
  - Include at least 1, at most 5 evidences.
  - Each evidence must include evidence\_start\_time, evidence\_end\_time, evidence\_instance\_id, and a concise evidence\_rationale.
- Distractor
  - Include exactly one distractor option. Do not include any instance\_id.
  - Match the length and complexity of answer\_text to avoid guess-by-length.
\newline

CONSTRAINTS:

- Question
  - Do not include any instance\_id. Make the subject uniquely identifiable via appearance, actions, location, and timing descriptions.
  - Questions must have human as the subject; phrase the question so the human is the grammatical subject. Do not use non-human entities (objects, devices) as the subject in the question.
  - Avoid generic questions that could be answered by observing cues common to most instances in the scene; prefer questions that must depend on a small subset of specific evidences to be answered correctly.
  - About the small subset of specific evidences, prefer questions that require multiple different instances to answer, rather than those that can be answered using only a single instance. i.e., we prefer questions that should be answered by observing multiple evidences.
  - Questions must be answerable solely from the video; the test taker will not have access to the Combined Video Descriptions.
  - The question must be extremely difficult for an AI model to answer correctly without genuinely understanding the nuanced details of the video context.
- Answer
  - Answer should be concise and uniquely correct. within 15 words. Do not include any instance\_id.
- Evidence
  - evidence\_instance\_id may refer to either a human or an object that supports the reasoning.
  - Make each evidence decisive and necessary: if that visual evidence were occluded or removed from the video, the question would become unanswerable or the answer would no longer be uniquely determinable.
  - Choose a minimal, non-redundant set of decisive evidences; avoid including evidence that is merely suggestive or duplicative.
- Distractors
  - Ensure the distractor option is incorrect but plausible; mutually exclusive with the correct answer and contradict the video/descriptions.
  - You may create distractors basing on another salient instance in the video or commonsense expectations about the scene, such as inventing a cause or event that sounds logical or physically possible but did not happen in the video.
  - The distractor MUST mimic the linguistic structure, length, and level of detail of the Correct Answer, so that it is extremely difficult for an AI model to choose correctly.
- General
  - Ground all content strictly in the provided descriptions/video; do not hallucinate unseen entities, times, or actions.
  - Do not alter, rename, or reassign any instance\_id. Use instance IDs exactly as defined in the Combined Video Descriptions for both subject\_instance\_id and evidence\_instance\_id to ensure traceability and alignment.
\newline

OUTPUT REQUIREMENTS:
- Single, valid JSON object that exactly matches the QA-Evidence Template. No prose before or after the JSON. No extra commentary or sentences.
\newline

FINAL VALIDATION STEP 1: Try to answer each question without watching the video:
After generating the QA, please try to answer the question without watching the video. If you can answer correctly with only common sense reasoning but without any information from the video, please regenerate the QA because the original QA is too easy to answer.

FINAL VALIDATION STEP 2: Watch the Video to validate each generated QA-Evidence item. Confirm that:
- The question subjects and evidence instances correspond to the correct pixel regions and entities observed in the video.
- The causal relationships implied in the question and answer are visually justified — i.e., the cause and effect are temporally and spatially consistent.
- The Answer is unique and unambiguous, directly supported by the cited evidences without conflicting viusal cues elsewhere.
- The distractor option is clearly incorrect yet plausible, ensuring it could mislead someone who has not fully reasoned through the causal sequence.
If any of these checks fail, regenerate or revise that QA-Evidence item until all conditions are fully satisfied.

FINAL VALIDATION STEP 3: (before returning):
- JSON parses; no code fences; no placeholders remain.
- question\_text and answer\_text contains no instance\_id; subjects are uniquely identifiable from wording and timing.
- All instance\_ids exactly match the IDs in Combined Video Descriptions; none are renamed or re-assigned.
\end{tcolorbox}
\end{figure*}

\begin{figure*}
\label{prompt5:qa_evidence_template}
\begin{tcolorbox}[
    colback=gray!15,
    colframe=gray!50,
    enhanced,
    arc=6pt,
    boxrule=1pt,
    title={Prompt 5: QA\_Evidence Template},
    fonttitle=\bfseries,
    colbacktitle=gray!40,
    coltitle=black,
    left=5pt,right=5pt,top=5pt,bottom=5pt
    ]
\scriptsize
\begin{verbatim}
{
    "QA_Evidence_List": [
      {
        "Question": {
          "question_text": "[Question text placeholder]",
          "subject_instance_id": "[Instance ID placeholder]",
          "question_start_time": "[Time placeholder]",
          "question_end_time": "[Time placeholder]",
          "question_type_subcategory": "[Number placeholder]",
          "question_type_subcategory_name": "[Category name placeholder]",
          "video_scene_category": "[Scene category placeholder]"
        },
        "Answer": {
          "answer_text": "[Answer text placeholder]"
        },
        "Evidences": [
          {
            "evidence_start_time": "[Time placeholder]",
            "evidence_end_time": "[Time placeholder]",
            "evidence_instance_id": "[Instance ID placeholder]",
            "evidence_rationale": "[Rationale text placeholder]"
          },
          (up to 5 evidence items like the first one)
        ],
        "Distractor": {
            "distractor_option_text": "[Distractor option text placeholder]"
        }
      },
      (totally 6 QA-Evidence items like the first one)
    ]
}
\end{verbatim}
\end{tcolorbox}
\end{figure*}

\begin{figure*}[b]
\label{prompt6:Causal_Question_Taxonomy}
\begin{tcolorbox}[
    colback=gray!15,
    colframe=gray!50,
    enhanced,
    arc=6pt,
    boxrule=1pt,
    title={Prompt 6: Causal Question Taxonomy},
    fonttitle=\bfseries,
    colbacktitle=gray!40,
    coltitle=black,
    left=5pt,right=5pt,top=5pt,bottom=5pt
    ]
\scriptsize
\begin{verbatim}
## CATEGORY 1: CAUSAL EXPLANATION
**Purpose:** Questions that explain the reasons (why) or mechanisms (how) behind actions or events, 
based on spatiotemporal evidence.

### Subcategory 1.1: Why Questions (Causal Reasons)
**Description:** These questions seek to understand the underlying causes and motivations behind ob-
served actions or events. They require reasoning about causal chains and identifying the root causes 
that led to specific outcomes.
**Causes may include:**
- Direct triggers (immediate causes that set events in motion)
- Enabling conditions (circumstances that made the action possible)
- Mediating mechanisms (intermediate processes that connect cause to effect)
- Rule-based causality (social, legal, or procedural reasons)

### Subcategory 1.2: How Questions (Mechanisms)
**Description:** These questions focus on understanding the processes, methods, and mechanisms throu-
gh which specific outcomes were achieved. They require analyzing the step-by-step procedures and tec-
hniques used.
**Mechanisms may include:**
- Sequential steps (ordered procedures and workflows)
- Tool use (utilization of instruments and equipment)
- Exploitation of constraints/supports (leveraging environmental or structural elements)
- Path/obstacle handling (navigation and problem-solving strategies)
- Socially/rule-driven processes (following protocols and social conventions)

## CATEGORY 2: COUNTERFACTUAL REASONING
**Purpose:** Questions that test the ability to deduce the most direct and unavoidable consequence of 
a single, precisely defined alteration to a key causal element within the scene, while holding all o-
ther scene variables constant.
**Mechanism for Uniqueness**: The question must isolate a single variable and propose a minimal, con-
crete change. The correct answer is the primary effect that follows directly from this change accord-
ing to physical laws, established procedures, or strong social norms, before any secondary reactions
or adaptations can occur.

### Subcategory 2.1: Physical Counterfactual
**Description:** Questions that explore the immediate physical outcome after altering a single mater-
ial property, environmental condition, or piece of equipment. The change must be specific enough to 
trigger a deterministic outcome based on physics (e.g., gravity, momentum, structural failure).
\end{verbatim}
\end{tcolorbox}
\end{figure*}

\begin{figure*}
\begin{tcolorbox}[
    colback=gray!15,
    colframe=gray!50,
    enhanced,
    arc=6pt,
    boxrule=1pt,
    title={Causal Question Taxonomy (Continued)},
    fonttitle=\bfseries,
    colbacktitle=gray!40,
    coltitle=black,
    left=5pt,right=5pt,top=5pt,bottom=5pt
    ]
\scriptsize
\begin{verbatim}
- Equipment failure (malfunction of tools, devices, or systems)
- Environmental alteration (changes in weather, lighting, terrain, or surroundings)
- Removal of supports/constraints (elimination of structural or physical aids)
- Parameter changes (modifications to speed, strength, size, or other physical properties)
- Alternative actions (different physical approaches or methods)



### Subcategory 2.2: Social Counterfactual
**Description:** Questions that explore the immediate social outcome after altering a single, critical
social variable—such as a specific instruction, the removal of a person with a unique role, or a viol-
ation of a clear protocol. The outcome should be the most direct consequence based on the established 
social or rule-based dynamics.
- Absence of participants (removal of key people from the scenario)
- Behavioral shifts (changes in individual or group behavior patterns)
- Altered instructions or rules (modifications to guidelines, protocols, or social norms)
- Communication differences (changes in how people interact and exchange information)
- Group dynamics changes (shifts in team composition, leadership, or social hierarchy)

---

## CATEGORY 3: PREDICTIVE ANTICIPATION
**Purpose:** To generate questions that require predicting the most probable and immediate outcome or 
next action in a sequence, based on a clearly established trajectory (physical, intentional, or proc-
edural) derived from unambiguous spatiotemporal evidence.
**Mechanism for Uniqueness**: The question must be based on a scene where actions are already in mot-
ion or a clear preparatory state has been established. The prediction should be limited to the logic-
al continuation or completion of this ongoing action within the next moment (e.g., 1-2 seconds), the-
reby excluding more distant or speculative possibilities.

### Subcategory 3.1: Physical Prediction
**Description:** Questions that predict the immediate future state of an object or system based on 
its current motion, trajectory, and interaction with the immediate environment. The scenario must 
provide sufficient physical information (e.g., velocity, direction, position relative to obstacles) 
to make the next state highly deterministic.
**Focus:** Predicts outcomes based on physical laws:
- Motion trajectory (path and velocity of moving objects)
- Stability (balance, structural integrity, and equilibrium states)
- Collisions (interactions between objects and their consequences)
- Environmental effects (impact of weather, terrain, and external conditions)
- Threshold states (critical points where systems change behavior)

### Subcategory 3.2: Behavioral Prediction
**Description:** Questions that predict a person's next immediate action based on a sequence of beha-
vior that clearly indicates a short-term goal or intention. The evidence must include preparatory ac-
tions, focus of attention (gaze), and body posture that collectively point to a single, imminent action.
**Focus:** Predicts human actions based on:
- Behavioral patterns (consistent ways individuals or groups typically act)
- Habits (routine behaviors and automatic responses)
- Social norms (expected behaviors within cultural or social contexts)
- Immediate intentions (short-term goals and immediate objectives)
- Interaction cues (non-verbal signals and social indicators)
Note: This subcategory inherently focuses on humans; ensure animals/human-operated objects are included
 only if tied to the actor’s behavior.

## CATEGORY 4: INFERENTIAL DESCRIPTION
**Purpose:** Questions that infer implicit attributes or states that are not directly visible or expli-
citly stated, grounded in multiple observable cues (not mere description).

### Subcategory 4.1: Identity/Role Inference
**Description:** These questions require deducing a person's professional role, social status, or iden-
tity based on observable indicators rather than explicit information. They involve pattern recognition 
and social knowledge.
**Description:** Infers a person's role or profession from:
- Clothing (uniforms, professional attire, protective gear, or status indicators)
- Tools (specialized equipment, instruments, or devices characteristic of specific roles)
- Directive behavior (leadership actions, supervisory activities, or authoritative gestures)
- Others' responses (following instructions, seeking help, coordinating around the person)

### Subcategory 4.2: Interpersonal Relationship Inference
**Description:** Determine the relationship type between people (e.g., colleagues, supervisor-subordin-
ate, caregiving/affectional) based on interaction style, proximity, and collaboration patterns.
**Evidence cues:**
- Distance and touch (intimate contact, protective stance, personal space)
- Collaboration and division of labor (signs of ongoing coordination, complementary tasks)
- Communication style (formal vs. informal, directive vs. consultative)
- Context and setting (home/work/public-service role expectations)

\end{verbatim}
\end{tcolorbox}
\end{figure*}

\begin{figure*}
\begin{tcolorbox}[
    colback=gray!15,
    colframe=gray!50,
    enhanced,
    arc=6pt,
    boxrule=1pt,
    title={Causal Question Taxonomy (Continued)},
    fonttitle=\bfseries,
    colbacktitle=gray!40,
    coltitle=black,
    left=5pt,right=5pt,top=5pt,bottom=5pt
    ]
\scriptsize
\begin{verbatim}

### Subcategory 4.3: Goal/Intent Inference (non-immediate)
**Description:** Infer a person's short-term goal or task state without predicting the next second of 
action (to avoid overlap with predictive categories).
**Evidence cues:**
- Preparation and planning (arranging/checking tools, tuning equipment, consulting a list)
- Attention and path (gaze target, heading, pointing gestures)
- Rule/procedure alignment (standard layouts, adherence to task steps)

### Subcategory 4.4: Emotion/Motivation Inference
**Description:** Infer likely emotional state or underlying motivation (e.g., tense, relaxed, urgent, 
helpful) from facial cues, action tempo, and situational context.
**Evidence cues:**
- Face and eyes (muscle tension, avoidance vs. steady gaze)
- Body tempo (hurried vs. slow, hesitation, force modulation)
- Situational events (setbacks, threats, recent completion of a goal)

### Subcategory 4.5: Skill/Expertise Inference
**Description:** These questions involve assessing a person's level of competence, expertise, or profe-
ssional skill based on their performance quality and behavioral indicators. They require understanding 
of skill development and professional standards.
**Focus:** Infers level of skill or professionalism from:
- Fluency (smoothness and efficiency of task execution)
- Correctness (accuracy and adherence to proper procedures)
- Error rates (frequency and types of mistakes made)
- Precision in performance (attention to detail and quality of work)
- Confidence indicators (certainty, hesitation, and self-assurance)
- Others' responses (seeking help vs. giving guidance)

### Subcategory 4.6: Ownership/Belonging Inference
**Description:** These questions involve determining who owns or has rights to specific objects, spaces, 
or resources based on behavioral patterns and spatial relationships. They require understanding of owner-
ship indicators and territorial behavior.
**Focus:** Infers ownership of objects based on:
- Placement (where objects are positioned relative to people)
- Guarding (protective behaviors and territorial actions)
- Frequency of use (how often and how naturally someone interacts with objects)
- Access patterns (who has permission to use or modify items)
- Personalization (customization and individual modifications)

### Subcategory 4.7: Risk/Safety Assessment
**Description:** These questions involve evaluating potential dangers, safety levels, and risk factors 
in a given situation. They require understanding of safety indicators, hazard recognition, and risk as-
sessment principles.

**Focus:** Infers level of hazard from:
- Unstable structures (structural integrity and potential collapse points, heat, sharp edges, fall 
risk, load issues)
- Warning signs (safety indicators, barriers, and cautionary signals)
- Crowd and individual reactions (avoidance, alarm, cautious handling)
- Environmental conditions (weather, lighting, terrain, and external threats)
- Safety equipment (presence or absence of protective gear and safety measures)

### Subcategory 4.8: Success/Failure Inference
**Description:** These questions involve determining whether an action, task, or process is succeed-
ing or failing based on observable outcomes, behavioral reactions, and performance indicators. They 
require understanding of success markers and failure signals.
**Focus:** Infers success or failure from:
- Outcome indicators (visible results and achievement markers)
- Behavioral reactions (satisfaction, frustration, celebration, or disappointment)
- Performance quality (accuracy, efficiency, and adherence to standards)

### Subcategory 4.9: Environmental Condition Inference
**Description:** Infer weather/temperature/surface/lighting or similar environmental conditions from 
human adaptations, equipment, and environmental traces.
**Evidence cues:**
- Adaptive behavior (seeking cover, tightening clothing, squinting, slow gait)
- Equipment usage (umbrellas, flashlights, hand warmers, crampons)
- Surface traces (puddles, reflections, dust, ice)
- Lighting and visibility (glare, backlight, dimness, fog)

### Subcategory 4.0: Other Contextual Inference
**Description:** Infer factors not covered above but relevant to understanding the scene, such as 
time of day, culture/customs, economic/resources, or rules/regulations.
**Evidence cues:**
- Time cues (shadow length, store open/closed state, festive decorations)
- Cultural/customary markers (attire styles, rituals, language/iconography)
- Economic/resource signs (equipment age, material sufficiency)
- Rules/regulation (signage, procedures, checkpoints)
\end{verbatim}
\end{tcolorbox}
\end{figure*}

\subsection{Multiple-Choice Question Generation}
\label{sec:supp_pipeline_multiple_choice}

To create a rigorous multiple-choice evaluation format, we developed a sophisticated, multi-stage pipeline to generate a set of plausible yet incorrect options (distractors). The primary challenge is to design distractors that probe for genuine causal reasoning, making it difficult for models to arrive at the correct answer by relying on superficial linguistic patterns or visual biases. Our approach is motivated by the need to mitigate the influence of confounders, as illustrated in \cref{fig:causal_graph}, ensuring that a correct answer is selected based on a sound understanding of the causal chain of events in the video rather than spurious correlations. The process involves generating two main types of distractors, text-based and video-based, and then further refining them to create subtle variations that test the precision of a model's understanding. The generation process unfolds in three main steps:

\paragraph{Step 1: Text-Based Distractor Generation}
The first step is to create a distractor based only on the textual content of the question. Without access to the video, a large language model is prompted to generate a plausible alternative answer. The model is instructed to create an answer that is fundamentally different from the correct one, relying on common sense and the context provided in the question alone. The prompt emphasizes that the generated answer must not introduce new objects, people, or elements that are not mentioned in the question, and it must be concise (within 15 words) to match the style of the correct answer. This technique is designed to produce options that appeal to models with strong language priors but weak visual grounding, effectively testing for over-reliance on linguistic cues. If a model selects this distractor, it indicates that the model is relying on spurious linguistic correlations rather than genuine visual evidence from the video.

\paragraph{Step 2: Verification of Semantic Exclusivity}
After generating a text-based distractor, we ensure that it is semantically distinct from both the correct answer and the pre-existing video-based distractor (which is created from causally irrelevant objects in the video, as described in Step 3 of the QA generation process). Another language model is prompted to compare pairs of options and confirm that they do not carry the same meaning. The verification prompt asks the model to determine whether two answers have the same meaning, requiring a binary response. If a generated distractor is found to be a mere paraphrase or synonym of another option, it is discarded, and the generation process is repeated up to a maximum number of attempts. This verification step is crucial for maintaining a set of genuinely different choices, as semantically equivalent options would not provide meaningful discrimination between models with different levels of understanding.

\paragraph{Step 3: Near-Miss Option Generation}
To further increase the difficulty and test the precision of a model's understanding, we generate ``near-miss'' versions for each of the three options: the correct answer, the text-based distractor, and the video-based distractor. A language model is tasked with creating a new version of each option that adheres to the same sentence structure and template but alters exactly one key attribute. The prompt specifies that each near-miss option must match the length within a small tolerance (approximately 10\% of the original) and that all six resulting options (three originals plus three near-misses) must be mutually exclusive. This creates a set of highly similar pairs of options, where only one is correct. This forces the model to perform fine-grained reasoning and pay close attention to critical details, as a superficial understanding would be insufficient to distinguish between the correct option and its near-miss counterpart. The near-miss design is particularly effective at testing whether models can identify subtle but decisive differences in causal reasoning chains.

Through this pipeline, each question is equipped with a comprehensive set of six options: the correct answer, a text-based distractor, a video-based distractor, and a near-miss version for each. This design enables a fine-grained analysis of model failures, distinguishing between errors caused by linguistic biases (selection of text-based distractors), visual misinterpretations (selection of video-based distractors), or a lack of detailed causal understanding (selection of near-miss options). The multi-layered distractor strategy ensures that achieving high accuracy on CaST-Bench requires models to construct precise, grounded causal chains rather than relying on surface-level cues.

\subsection{Mask-Based QA Filtering for Causal Chain Validation}
\label{sec:supp_pipeline_mask_based}

After generating QA pairs with causal chains, we must ensure that questions genuinely require the visual evidence we provide and cannot be answered through spurious correlations or incomplete reasoning chains. As discussed in \cref{sec:cast_bench}, mitigating the influence of confounders is fundamental to causal inference, and our filtering procedure addresses this by validating both the necessity of visual evidence and the completeness of the causal chain. We employ a three-step filtering process: first, a text-based filter removes questions answerable without video access; second, a novel video-masked filter validates that questions require the specific spatio-temporal evidence we provide; and finally, human annotators conduct thorough review to ensure quality.

\paragraph{Step 1: Text-Based Filtering}
Some questions may be answerable through linguistic patterns alone without requiring visual evidence, allowing models to rely on spurious linguistic correlations rather than genuine visual understanding. To ensure questions require actual video evidence, we present the question and all answer options to a language model without providing the video. If the model successfully identifies the correct answer, this indicates the question can be solved through text-only reasoning and is filtered out.

\paragraph{Step 2: Video-Masked Filtering}
Even when questions pass the text-based filter, they may still be answerable through unintended visual cues, dataset biases, or incomplete causal chains that miss critical evidence. The challenge here is to validate that the causal chain we provide is complete --- that is, \textit{questions should truly require the specific spatio-temporal evidence we identify, and removing this evidence should make the question unanswerable}. This validation is motivated by the need to ensure that our benchmark tests genuine causal reasoning rather than superficial pattern matching or reliance on unintended visual context.

To address this, we implement a video-masked filtering procedure. First, we collect all bounding box regions associated with both the question subject (if specified) and all evidence instances from the causal chain. These regions represent the spatio-temporal evidence that should be necessary to answer the question. We then create a masked version of the video where these evidence regions are blacked out by setting the corresponding pixels to black.

The masked video is then presented along with the question and all answer options to a vision-language model. The model is instructed to answer based only on visible evidence from the video and to select a special option indicating ``Unable to determine'' if insufficient information is available to make a confident choice. If the model can still correctly answer the question using the masked video, this indicates one of three problematic scenarios: (1) the causal chain is incomplete, missing critical evidence that would be necessary for a human to answer correctly; (2) the question can be solved through unintended visual cues outside the masked regions, such as background context or other objects that happen to be visible; or (3) dataset biases or question wording allow correct answers without the intended evidence. Questions that pass this masked-video test are filtered out, ensuring that only questions requiring the complete, intended causal chain remain in the benchmark.

After these automated filtering steps, human annotators conduct a final review and revise to validate the question, answer, and causal chain, ensuring that the remaining questions meet our quality standards. Through this rigorous three-step filtering process, only approximately 40\% of the initially generated QA pairs remain, as reported in \cref{sec:cast_bench}, resulting in a benchmark that truly tests causal reasoning grounded in spatio-temporal evidence.

\section{Details of Experiment Setup}
\label{sec:supp_experiment_setup}

\subsection{Evaluation Prompt}
\label{sec:supp_experiment_setup_evaluation_prompt}

All evaluated VLMs shared a single unified prompt for video QA. For the multiple-choice setting, the exact prompt is provided in \hyperref[prompt7:cast_bench_evaluation_prompt]{Prompt 7}.

\subsection{VLM Hyperparameter Configuration}

We configure all VLMs with \texttt{max\_new\_tokens} set to 2048, limiting each sample to at most 2{,}048 generated tokens (excluding the input length), and \texttt{cutoff\_len} set to 204{,}800, which caps the combined input-plus-output length at 204{,}800 tokens so that truncation is effectively disabled and long contexts are preserved end-to-end.

\begin{figure*}
\begin{tcolorbox}[
    colback=gray!15,
    colframe=gray!50,
    enhanced,
    arc=6pt,
    boxrule=1pt,
    title={Prompt 7: CaST-Bench Evaluation Prompt},
    fonttitle=\bfseries,
    colbacktitle=gray!40,
    coltitle=black,
    left=5pt,right=5pt,top=5pt,bottom=5pt
    ]
\scriptsize
\label{prompt7:cast_bench_evaluation_prompt}
\begin{verbatim}

<video> {question}

Please analyze the video and answer the multiple-choice question by selecting the most appropriate option, and
provide the visual evidences from the video that support your answer.

Your response should be in the following JSON format (Do not include ellipses; Ensure the JSON is self-contained
and valid):

OUTPUT TEMPLATE (replace with your content; do not include placeholders):
{
  "instances": [
    {
      "instance_name": "short name of the instance",
      "evidences": [
        {
          "evidence_start_time": "mm:ss",
          "evidence_end_time": "mm:ss",
          "evidence_rationale": "evidence description",
          "bboxes_in_time_range": {
            "ss": "[x_min, y_min, x_max, y_max]",
            "ss": "[x_min, y_min, x_max, y_max]",
            "ss": "[x_min, y_min, x_max, y_max]",
            ...
          }
        },
        ...
      ]
    },
    ...
  ],
  "answer_choice": "the chosen option"
}

STRICT OUTPUT ONLY
- Output a single valid JSON object. No prose, no code fences, no comments.
- Use double quotes for all strings. No trailing commas.
- Keep key names exactly as specified. Do not add, remove, or rename keys.
- Keep the key order as shown in the template.

SCHEMA AND CONSTRAINTS
- "instances": array of one or more instances, each with:
  - "instance_name": short noun phrase identifying the entity (string).
  - "evidences": array of one or more evidences for that instance.
- Each evidence must have:
  - "evidence_start_time": start timestamp in "mm:ss" (zero-padded).
  - "evidence_end_time": end timestamp in "mm:ss" (>= start, zero-padded).
  - "evidence_rationale": one concise sentence explaining this evidence clip.
  - "bboxes_in_time_range": object whose keys are every whole-second timestamp from "evidence_start_time" to
  "evidence_end_time" (inclusive) in "mm:ss" format (zero-padded), and whose values are strings formatted exactly 
  as "[x_min, y_min, x_max, y_max]" with integers. No missing seconds.
- The sum of all evidences across all instances must be <= 5.
- "answer_choice": the single uppercase option letter that best matches the answer.

\end{verbatim}
\end{tcolorbox}
\end{figure*}

\section{Evaluation Suite}
\label{sec:supp_evaluation_suite}

\subsection{Grounded Causal Chain Evaluation}
\label{sec:supp_evaluation_suite_grounded_causal}

Evaluating the correctness of a predicted causal chain is fundamentally harder than checking a single grounding target. A model must recover every actor that participates in the causal process, align their evidences across different time ranges, and ensure the supporting boxes stay faithful to the underlying video. Because the number and ordering of predicted instances rarely match the ground truth, we design a three-stage procedure: (i) greedy instance matching, (ii) temporal accuracy scoring, and (iii) spatio-temporal accuracy scoring. The latter two stages use the IM-tIoU and IM-vIoU metrics defined in \cref{sec:eval} in the main paper, so we expand here on the instance matching procedure that enables those metrics.

\paragraph{Instance Matching}
The central difficulty lies in linking each predicted instance to a unique ground-truth counterpart without relying on shared identifiers. A naive name-based match fails because models may rename entities, merge multiple actors, or hallucinate new ones. Instead, we compare every predicted--ground-truth pair using a spatio-temporal overlap score that multiplies the temporal IoU between their evidence spans and the average spatial IoU of the aligned bounding boxes over the overlapping frames. This score naturally favors predictions that are both time-aligned and spatially localized. Once we obtain the score matrix between all predicted instances and GT instances, we apply a greedy selection process: the pair with the highest score is matched first, both elements are removed from further consideration, and the process repeats until no positive-overlap pairs remain. This one-to-one assignment ensures that each ground-truth instance contributes at most one match, preventing duplicated credit for the same evidence and allowing downstream metrics to treat unmatched predictions as false positives and missing instances as false negatives.

\paragraph{Temporal Accuracy (IM-tIoU)}
With the instance pairs fixed, we measure how accurately the model captured the timing of each evidence segment using the IM-tIoU metric introduced in \cref{sec:eval}. The metric averages temporal IoU scores across ground-truth instances, so any unmatched or poorly aligned instance directly lowers the final score.

\paragraph{Spatio-Temporal Accuracy (IM-vIoU)}
Finally, we apply the IM-vIoU metric from \cref{sec:eval} to assess joint spatial and temporal fidelity. Because the greedy matching step guarantees clean one-to-one assignments, IM-vIoU can focus purely on how well each matched pair overlaps in both time and space, rewarding predictions that retrieve the correct actors, durations, and bounding boxes across the entire causal chain.

\subsection{Open-Ended Evaluation}
\label{sec:supp_evaluation_suite_open_ended}

The quantitative metrics in \cref{sec:eval} diagnose how well a model grounds causal evidence and selects multiple-choice answers, yet they cannot fully capture the nuanced, free-form responses that arise when a system explains its reasoning without option prompts. To complement the automatic metrics, we introduce an open-ended evaluation protocol that enlists a large language model (LLM) as an expert judge. This protocol focuses on two challenges highlighted throughout CaST-Bench: (i) ensuring that causal chains remain faithful to the video even when they are written in natural language, and (ii) rewarding models whose final answers are logically supported by their own rationales rather than by dataset priors.


We prompt an LLM to act as an impartial evaluator. The full instruction provided to the judge is reproduced at \hyperref[prompt8:open_ended_evaluation]{Prompt 8}.

\begin{figure*}[t]
\begin{tcolorbox}[
    colback=gray!15,
    colframe=gray!50,
    enhanced,
    arc=6pt,
    boxrule=1pt,
    title={Prompt 8: Open-Ended Evaluation},
    fonttitle=\bfseries,
    colbacktitle=gray!40,
    coltitle=black,
    left=5pt,right=5pt,top=5pt,bottom=5pt
]
\label{prompt8:open_ended_evaluation}
\scriptsize
\begin{verbatim}
You are an expert evaluator of causal reasoning quality.

Given:
1.  A causal question about a video event {question}
2.  The ground-truth conclusion answer {gt_answer}
3.  The ground-truth causal reasoning process {gt_reasoning}
4.  A test-taker model’s generated conclusion answer {pred_answer}
5.  A test-taker model’s generated causal reasoning process {pred_reasoning}

Task:
Evaluate the test-taker model’s reasoning across four dimensions. Assign
a separate score (0–10) for each dimension according to the standards below.

Evaluation Dimensions:
1. Answer Conclusion Correctness
* Compare only the model’s conclusion answer to the ground-truth conclusion
  answer; ignore any reasoning content.
* Judge semantic equivalence, polarity, entity/attribute correctness, and
  numeric/unit consistency; penalize contradictions, material vagueness,
  or hedging that alters commitment.

2. Causal Chain Logical Consistency
* Evaluate only the generated answer and reasoning; do not reference ground-truth
  answer or ground-truth reasoning.
* Verify that causes precede effects and that the causal sequence is minimal yet
  sufficient to explain the answer.
* Identify any logical leaps, post-hoc reasoning, or teleological claims lacking
  justification.
* Ensure temporal feasibility and internal consistency across causal steps.

3. Evidence Coverage & Completeness
* Use the ground-truth causal reasoning as the reference. Check that the generated
  causal reasoning includes and aligns with its key entities, events, moments,
  and causal steps.
* Evaluate recall of essential causal steps and contextual conditions relative
  to the ground truth.
* Penalize missing core components, contradictions to the ground truth, or
  hallucinated steps; avoid rewarding overemphasis on partial evidence.

4. Evidence–Conclusion Overall Justification
* Consider the generated answer and the generated reasoning together: does the
  provided reasoning justify the stated answer, and do both align with the ground
  truth overall?
* Assess the logical coherence from evidence to conclusion, calibration of
  confidence, and global plausibility.

Scoring Standards for each dimension:
9~10: Exemplary performance with no flaws
7~8: Non-critical deviations present
5~6: Quality-impairing defects
3~4: Serious validity-compromising errors
0~2: Fundamental functionality failure

Output Format:
Output only a valid JSON object in the following format (no additional text):
{{"answer_conclusion_correctness": 00.00,
  "causal_chain_logical_consistency": 00.00,
  "evidence_coverage_completeness": 00.00,
  "evidence_conclusion_overall_justification": 00.00}}
\end{verbatim}
\end{tcolorbox}
\end{figure*}

\section{Case Studies and Failure Analysis}
\label{sec:supp_failure}




Beyond the quantitative ablation studies and error analysis presented in the main paper, we here perform a qualitative analysis of the case studies and failure patterns exhibited by the evaluated models.

\begin{figure*}
    \centering
    \includegraphics[width=0.85\linewidth]{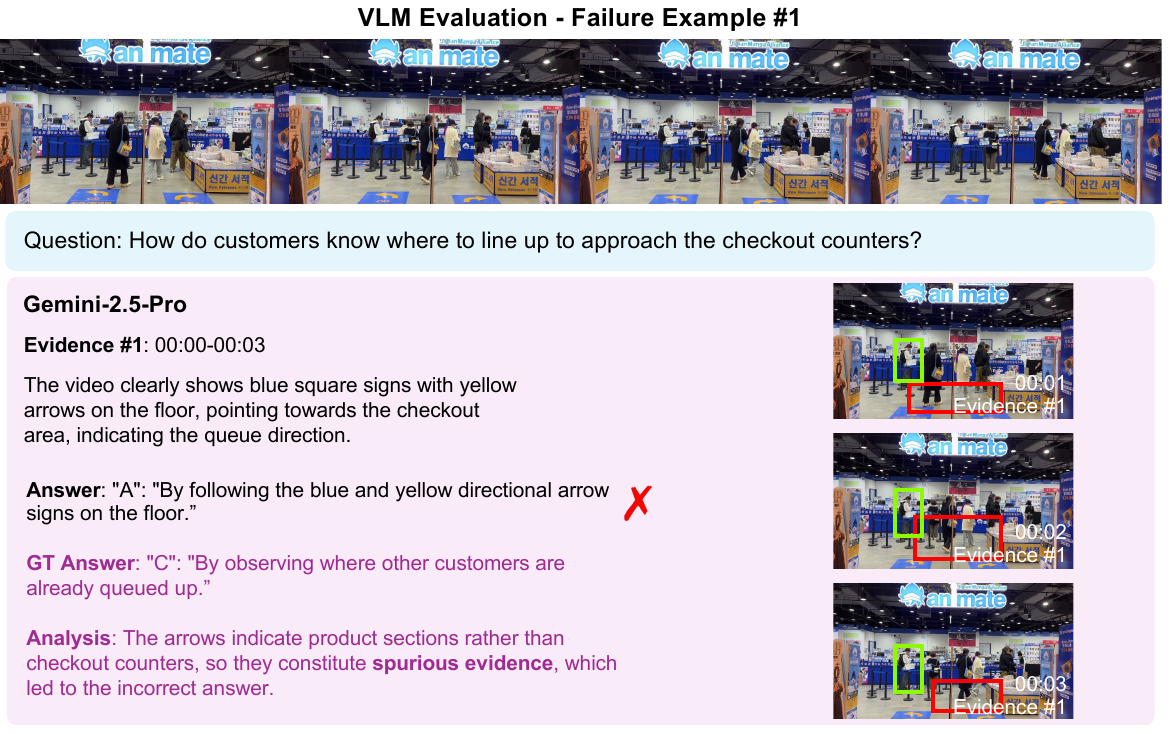}
    \caption{\textbf{Case Studies and Failure Analysis.} \#1 failure example.}
    \label{fig:supp_eval_sample2}
\end{figure*}

\paragraph{Vulnerability to Spurious Visual Confounders} A core design principle of CaST-Bench is the inclusion of distractors that target spurious correlations --- visual or linguistic cues that are salient but causally irrelevant. We observe that even top-tier proprietary models can be misled by these confounders when they fail to strictly follow the causal chain. As shown in the failure example of \cref{fig:supp_eval_sample2}, when asked \textit{``How do customers know where to line up?''}, Gemini-2.5-Pro \cite{comanici2025gemini} correctly identifies visual elements in the scene (blue and yellow arrows on the floor). However, it fails to distinguish between \textit{descriptive existence} (the arrows exist) and \textit{causal relevance} (the arrows indicate product sections, not queue lines). The model falls into the trap of the ``Video-Based Distractor'' (Option A), ignoring the true causal evidence --- the social cue of other customers already queuing (Option C). This error demonstrates that strong perception capabilities alone are insufficient; models must be able to filter out salient but non-causal visual information to reason correctly.

\begin{figure*}
    \centering
    \includegraphics[width=0.85\linewidth]{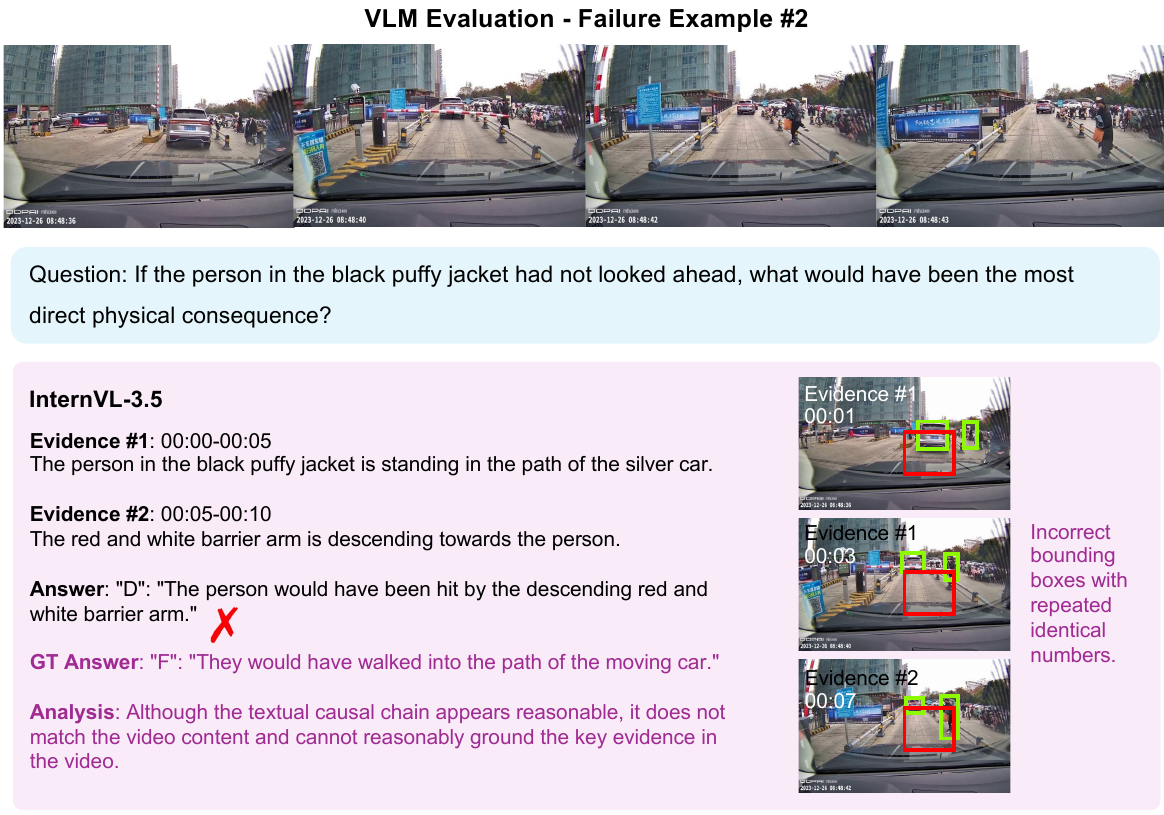}
    \caption{\textbf{Case Studies and Failure Analysis.} \#2 failure example.}
    \label{fig:supp_eval_sample3}
\end{figure*}

\paragraph{Visual Hallucination and Grounding Disconnect} Another significant failure pattern is the generation of plausible-sounding textual rationales that are completely disconnected from the actual visual data --- a phenomenon we term ``grounding disconnect''. In \cref{fig:supp_eval_sample3} (Failure Example \#2), InternVL-3.5 \cite{wang2025internvl3_5} attempts to answer a counterfactual question about a traffic scene. While the model selects a distractor answer (``hit by the descending barrier''), it attempts to justify this choice by hallucinating evidence. The generated bounding boxes for Evidence \#1 and \#2 are incorrect and repetitive, and the rationale incorrectly states that the barrier is moving downward toward the person, even though the video shows no such movement at those timestamps. This means a model may produce a logically coherent textual explanation that is physically grounded in hallucinated pixels, proving the necessity of enforcing explicit ST evidence grounding.

\begin{figure*}
    \centering
    \includegraphics[width=0.85\linewidth]{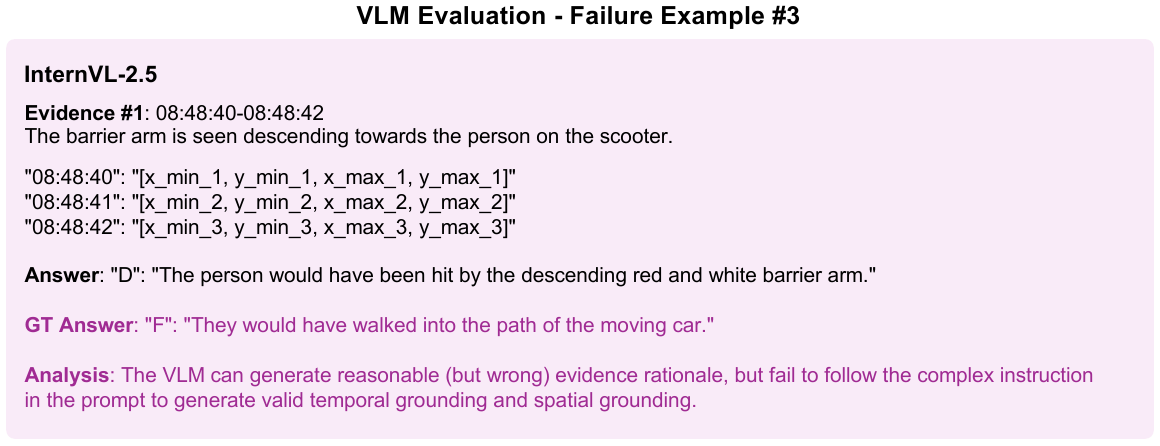}
    \caption{\textbf{Case Studies and Failure Analysis.} \#3 failure example.}
    \label{fig:supp_eval_sample4}
\end{figure*}

\paragraph{Instruction Following and Formatting Failures} The CaST-Bench task requires models to output a structured JSON containing precise timestamps and frame-by-frame bounding boxes. This poses a heavy instruction-following challenge. As illustrated in \cref{fig:supp_eval_sample4} (Failure Example \#3), InternVL-2.5 \cite{chen2024internvl25} successfully identifies the context of the question but fails to generate valid coordinates. Instead of calculating the specific bounding box values, the model simply repeats the placeholder templates (e.g., $[x_{min_1},...]$) provided in the prompt. This ``format collapse'' renders the output invalid for evaluation. This failure pattern suggests that complex spatio-temporal reasoning tasks require not only visual understanding but also robust instruction-following capabilities to map internal reasoning into precise, structured outputs.

\section{Social Impact, License, and Access}
\label{sec:supp_social}

\subsection{Broader Impact}

CaST-Bench advances the field of VLMs by shifting the focus from surface-level perception to deep, grounded causal reasoning, a capability essential for sophisticated video analysis and anticipation tasks. By mandating that models validate their answers with explicit spatio-temporal evidence, the benchmark significantly enhances transparency and trust, enabling the distinct identification of genuine understanding versus reliance on spurious correlations or hallucinations. Furthermore, the work introduces a scalable Human-AI collaborative pipeline for constructing high-quality, dense annotations, offering a methodological blueprint for future complex reasoning dataset creation. Ultimately, CaST-Bench serves as a rigorous diagnostic tool to mitigate visual confounders and biases, fostering the development of more robust, interpretable, and reliable multimodal AI systems.

\subsection{Limitations}
Although our human–AI collaborative pipeline yields high-quality spatio-temporal causal annotations, it limits scalability. Extending CaST-Bench to additional domains or substantially increasing its size would require retracing multiple stages of human verification to maintain quality. The benchmark inherits biases from the SegmentAnything-Video (SAV) corpus~\cite{ravi2024sam2}, such as the geographic distribution and activity types favored in that source dataset. Moreover, despite detailed annotation guidelines, human editors may introduce subtle preferences when rewriting rationales or validating causal chains, which could affect linguistic style or emphasis.

\subsection{Ethics, License, and Data Access}
All videos originate from the publicly released SAV dataset, and we follow the source licensing terms when curating clips. During data construction we filter out samples that contain sensitive or unsafe content and require annotators to follow institutional ethical standards, including anonymizing identifiable details beyond what is already visible in the original footage. We will release CaST-Bench (data, annotations, and evaluation code) under the Creative Commons Attribution 4.0 International (CC BY 4.0) license and the MIT License. This license grants broad research use provided that derived works cite the benchmark, while the underlying SAV assets remain governed by their original licenses and terms of use. Large multimodal models (e.g., Gemini-2.5-Pro) are used within the pipeline to bootstrap descriptions and candidate QA pairs; every machine-generated artifact is subsequently reviewed and revised by trained annotators to guarantee factual alignment.

\clearpage

\clearpage

\clearpage

\clearpage

\clearpage

\clearpage

\clearpage

\begin{figure*}
    \centering
    \includegraphics[width=0.85\linewidth]{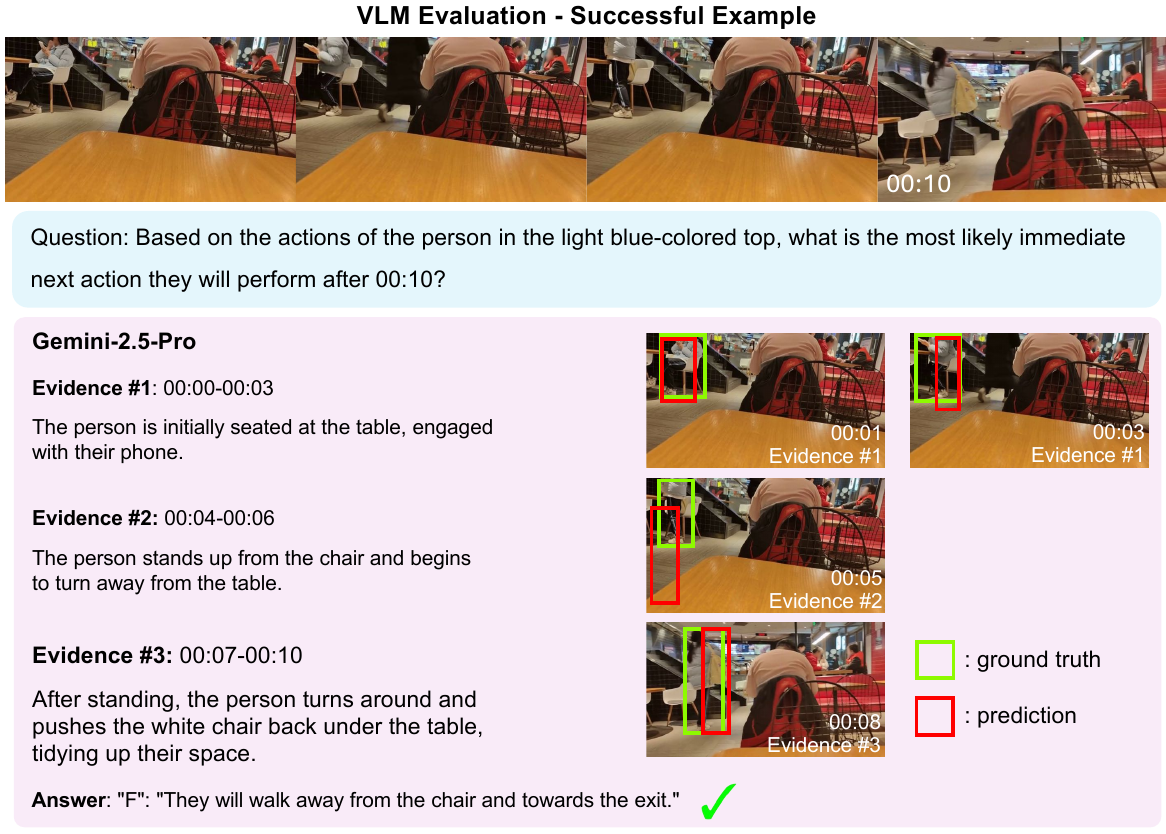}
    \caption{\textbf{Case Studies and Failure Analysis.} Successful example.}
    \label{fig:supp_eval_sample1}
\end{figure*}

\clearpage

\end{document}